%% file: main.tex
\documentclass{article}

    \usepackage[final]{neurips_2025}

\usepackage[utf8]{inputenc} % allow utf-8 input
\usepackage[T1]{fontenc}    % use 8-bit T1 fonts
\usepackage{hyperref}       % hyperlinks
\usepackage{url}            % simple URL typesetting
\usepackage{booktabs}       % professional-quality tables
\usepackage{amsfonts}       % blackboard math symbols
\usepackage{nicefrac}       % compact symbols for 1/2, etc.
\usepackage{microtype}      % microtypography
\usepackage{xcolor}         % colors
\usepackage{multirow}
\usepackage{graphicx}
\usepackage{wrapfig} 
\usepackage{colortbl}
\usepackage{pythonhighlight}
\usepackage{subfigure}
\usepackage[ruled,linesnumbered]{algorithm2e}

\title{%The direct preference optimization for score-based data assimilation\\
DAWP: A framework for global observation forecasting via \textbf{D}ata \textbf{A}ssimilation and \textbf{W}eather \textbf{P}rediction in satellite observation space 
}

\author{%
  Junchao Gong \thanks{Equal Contribution}\\
  Shanghai Jiao Tong University \\
  \texttt{gjchimself@sjtu.edu.cn} \\
  \And
  Jingyi Xu \footnotemark[1] \\
  Fudan University \\
  \texttt{jyxu22@m.fudan.edu.cn} \\
  \AND
  Ben Fei \thanks{Corresponding Authors: Ben Fei (benfei@cuhk.edu.hk) and Lei Bai (bailei@pjlab.org.cn)}  \\
  The Chinese University of Hong Kong \\
   Shanghai AI Laboratory \\
  \texttt{benfei@cuhk.edu.hk} \\
  \And
  Fenghua Ling \\
  Shanghai AI Laboratory \\
  \texttt{lingfenghua@pjlab.org.cn} \\
  \And
  Wenlong Zhang \\
  Shanghai AI Laboratory \\
  \texttt{zhangwenlong@pjlab.org.cn} \\
  \And
  Kun Chen \\
  Shanghai AI Laboratory \\
  \texttt{chenkun@pjlab.org.cn} \\
  \And
  Wanghan Xu \\
  Shanghai AI Laboratory \\
  \texttt{xuwanghan@pjlab.org.cn} \\
  \And
  Weidong Yang \\
  Fudan University \\
  \texttt{wdyang@fudan.edu.cn} \\
  \And
  Xiaokang Yang \\
  Shanghai Jiao Tong University \\
  \texttt{xkyang@sjtu.edu.cn} \\
  \And
  LEI BAI \footnotemark[2] \\
  Shanghai AI Laboratory \\
  \texttt{bailei@pjlab.org.cn} \\
}

\begin{document}

\maketitle

\begin{abstract}
 % weather prediction的重要性
 Weather prediction is a critical task for human society, where impressive progress has been made by training artificial intelligence weather prediction (AIWP) methods with reanalysis data.
 % AIWP 利用renalysis已经取得显著进展。但是reanalysis带给AIWP局限。
 % Impressive progress has been made by training artificial intelligence weather prediction (AIWP) methods with reanalysis data.
%  Taking the advantage of reanalysis data, artificial intelligence weather prediction (AIWP) methods achieve comparable or even better skills than 
%  leading numerical weather prediction (NWP) systems. 
 % However, the progress of producing reanalysis data limits the AIWPs with shortcomings such as temporal lag.
 % However, reliance on reanalysis data limits the AIWPs with shortcomings, such as temporal lag, hindering real-time forecasting.
 However, reliance on reanalysis data limits the AIWPs with shortcomings, including data assimilation biases and temporal discrepancies.
 % DOP的崛起
 % To overcome this constraint, observation forecasting emerges as a transformative paradigm, enabling AIWPs to operate independently of reanalysis data.
 To liberate AIWPs from the reanalysis data, observation forecasting emerges as a transformative paradigm for weather prediction.
 % DOP的核心问题之一high resolution observation data with variable missing values
 % One of the key challenges in observation forecasting is the irregular observation space constructed by high-resolution observation data with variable missing values, which constrains the design and prediction of AIWPs.
 One of the key challenges in observation forecasting is learning spatiotemporal dynamics across disparate measurement systems with irregular high-resolution observation data, which constrains the design and prediction of AIWPs.
 % 我们提出了observation space AIDA+AIWP的框架使DOP在一个规则的观测空间中进行
 To this end, we propose our DAWP as an innovative framework to enable AIWPs to operate in a complete observation space by initialization with an artificial intelligence data assimilation (AIDA) module.
 % 在AIDA中为了encoding irregulra observation data with multiple channels, 我们介绍了mask ViT-VAE.
 Specifically, our AIDA module applies a mask multi-modality autoencoder (MMAE) for assimilating irregular satellite observation tokens encoded by mask ViT-VAEs. 
 % 在AIWP中，我们介绍了CBC以进行基于sub-image的全球观测预报
 % For AIWP, we introduce a spatiotemporal decoupling transformer with cross-regional boundary conditioning (CBC) to enable sub-image-based global observation forecasting.
  For AIWP, we introduce a spatiotemporal decoupling transformer with cross-regional boundary conditioning (CBC), learning the dynamics in observation space, to enable sub-image-based global observation forecasting.
 % 我们进行了详细的实验验证了AIDA对于AIWP的roll out和efficient能力的提升。我们也展示了DOP具有应用于全球降水预报的潜力。
 Comprehensive experiments demonstrate that AIDA initialization significantly improves the roll-out and efficiency of AIWP. 
 Additionally, we show that DAWP holds promising potential to be applied in global precipitation forecasting. 
 Code will be available at this \href{https://github.com/jasong-ovo/DAWP-NIPS25}{github repo}.
\end{abstract}

\input{Sections/Introduction}

\input{Sections/Related_work}

\input{Sections/Method}

\input{Sections/Experiment}

\input{Sections/Conclusion}

\section*{Acknowledgements}
This work is Supported by Shanghai Artificial Intelligence Laboratory. This work was done during Junchao Gong's internship at Shanghai Artificial Intelligence Laboratory.

\bibliographystyle{unsrt}
\bibliography{ref}

\newpage
\clearpage
\input{checklist}

\newpage
\clearpage
\input{appendix}

\end{document}

%% file: Sections/Introduction.tex
\section{Introduction}
\label{sec:introduction}

% Global precipitation forecasting is crucial in weather prediction, as it influences a range of socioeconomic aspects, including transportation, agriculture, and public safety. 
Weather prediction is a critical task that significantly impacts various socioeconomic aspects, including transportation, agriculture, and public safety.
Traditional numerical weather prediction (NWP) systems rely on intricate human-designed workflows~\cite{dueben2018challenges,allen2025end}, such as numerical assimilation systems and physical solvers, to generate global precipitation predictions.

Recently, transformative progress has been made by artificial intelligence weather prediction (AIWP) models. 
These AIWP models now achieve forecast skill scores comparable to or even surpassing those of leading physics-based NWP systems~\cite{bi2023accurate,lam2023learning,chen2023fuxi,chen2023fengwu}. 
To learn the atmospheric dynamics, reanalysis products are widely used~\cite{hersbach2020era5,rienecker2011merra,gelaro2017modern,kalnay2018ncep}.

However, reanalysis data, generated by numerical data assimilation, introduce intrinsic limitations in AIWP models built upon them.
(I) \textbf{Data Assimilation Biases:} (Re)analysis products are synthesized by numerical data assimilation (DA), where direct observations are blended with a physics-based forecast. 
During DA, information loss of direct observation occurs due to the limited utilization of raw observational data and the preprocessing that resamples observations to regular grids of reanalysis data format with finite resolution~\cite{bauer2015quiet,valmassoi2023current,hu2023progress}. 
% , in which progress empirical blendings of direct observations with a traditional physics-based forecast are implemented.
% Information loss of direct observation can happen due to the limited data usage ratio and the preprocessing that resamples observations to regular grids of reanalysis data format with finite resolution~\cite{bauer2015quiet,valmassoi2023current,hu2023progress}. 
Additionally, the incomplete physical process modeling and uncertainty parameterizations in physics-based forecast systems also introduce biases, which could hinder learning the actual dynamics of the atmosphere~\cite{laloyaux2022deep}. 
% Biases can stem from incomplete physical process modeling and uncertainty parameterizations in physics-based forecast systems, which could hinder learning the actual dynamics of the atmosphere. 
% Information loss of direct observation can happen due to the limited data usage ratio and the resampling to regular grids of reanalysis data format with finite resolution. 
% (II) The temporal lag between direct observation acquisition and analysis data generation severely degrades the quick response ability of AIWP models to up to 6 hours~\cite{allen2025end,dong2025omg}.
(II) \textbf{Temporal Discrepancies:} 
The temporal lag between direct observation acquisition (nearly real-time) and analysis data generation (up to six hours) severely degrades the quick response ability of AIWP models~\cite{allen2025end,dong2025omg}.
These limitations may be further exacerbated by the discrepancy between real-world observation space and physical forecasting space which is required by NWP systems to implement dynamic equations of atmospheric~\cite{mcnally2024data}. 
As AIWP models do not require physical solvers to predict the evolution of the atmosphere, there is potential for them to directly predict atmosphere states in real-world observation space.

\begin{wrapfigure}{t}{0.5\columnwidth}
    \centering
    \includegraphics[width=0.5\columnwidth]{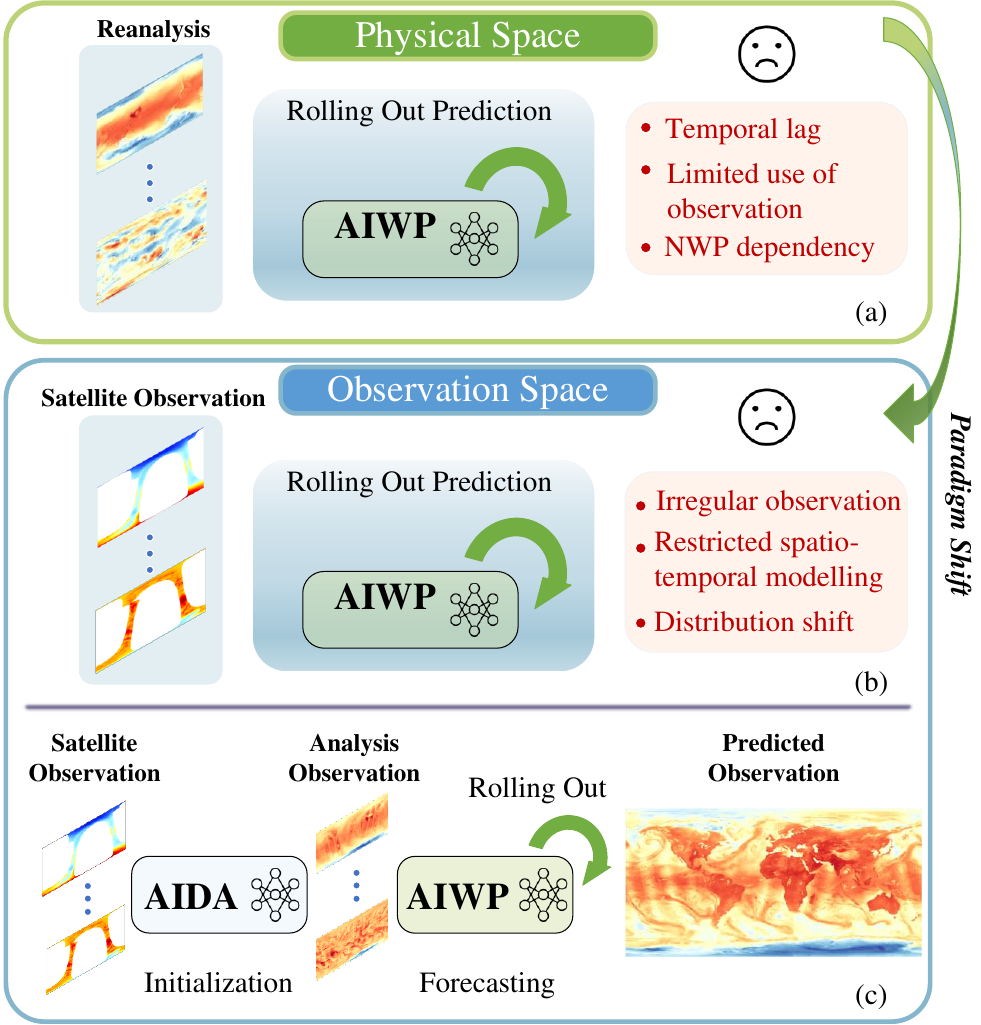}
    \vspace{-0.7cm}
    \caption{A paradigm shift from physical space to observation space. Our DAWP is illustrated in (c).}
    \label{fig:teaser}
\end{wrapfigure}

Artificial Intelligence Direct Observation Prediction (AI-DOP) is emerging as a transformative, data-driven approach with the potential to overcome the limitations in reanalysis-driven AIWP methods. 
The key challenge of AI-DOP is learning the spatiotemporal dynamics not only within a single observation source but also across disparate measurement systems, given the irregular and high-resolution observation data~\cite{mcnally2024data}.  
The spatiotemporal modeling approaches are restricted to learn the relationships between different observations with irregular data. ~\cite{mcnally2024data} applies a transformer with mask tokens to reformat multiple observations into regular ones.
Further, ~\cite{alexe2024graphdop} uses a graph encoder to flexibly encode different measurements into a uniform latent representation for latent forecasting with a naive transformer backbone.
In addition, as AI-DOP requires being generalized to any location grids given observations with variable missing values~\cite{mcnally2024data}, the dense forecasts and sparse observation inputs result in \textbf{input-output distribution shift} in rollout as shown in (b) and (c) of Figure~\ref{fig:teaser}.
Motivated by these questions, we argue that training AIWP models in \textbf{a uniform observation space} where input and output are both regular grid data.

We propose our DAWP, an AI-DOP system composed of an observation space data assimilation (AIDA) module and an AIWP module, to learn the spatiotemporal dynamics between various satellite observations with irregular and high-resolution characteristics.
% , and map observation predictions into precipitation. 
To begin with, a transformer VAE encoder/decoder is designed with observation masks to regionally encode high-resolution irregular direct earth observations for efficient I/O and computation.
Then, we implement a sub-image-based AIDA module for observation space data assimilation by a multi-modal masked autoencoder with encoded observation tokens. 
By learning the spatiotemporal correlations between various observations in the AIDA initialization, irregular observations are transformed into a uniform completed observation space.
In this imputed space, we train our sub-image-based AIWP module with Cross-regional Boundary Conditioning (CBC), which could forecast observations with a global state cache providing atmospheric states of neighbours.
Finally, a combination of mapping operators is used to obtain global observation predictions or precipitation variables.  
We conduct comprehensive experiments to demonstrate the effectiveness of our DAWP framework for global direct observation predictions.
We summarize the contributions of this paper as follows:
\begin{itemize}
\vspace{-0.2cm}
\setlength{\itemsep}{0pt}
\setlength{\parskip}{0pt}
    \item \textbf{Innovative framework} integrating AIDA and AIWP methods for direct observation predictions: We propose a brand-new framework that leverages an observation space AIDA module, transforming irregular observations into a uniform observation space, with AIWP modules to achieve skillful direct observation predictions, bypassing the limitations of reanalysis data.
    \item \textbf{High-resolution global forecasting} for observation and precipitation: We introduce a mask ViT-VAE and a spatiotemporal transformer with cross-regional boundary information for encoding high-resolution irregular observations, and implement global forecasting efficiently on sub-images of global observations.
    \item \textbf{Comprehensive experiments}: We organize a composite satellite observation dataset with a size of over 35TB, which has a spatiotemporal resolution of 12$\times$1152$\times$2304. 
    Comprehensive experiments and reanalyses are presented, demonstrating the effectiveness of our DAWP framework and the potential of direct observation predictions. 
\end{itemize}

%% file: Sections/Related_work.tex
\section{Related work}
\label{sec:related}

\textbf{Weather prediction with deep learning.}
% weather prediction's comparable ability
Recent studies have demonstrated that machine learning systems can produce accurate medium-range forecasts, comparable to physics-based models, for key weather parameters~\cite{bi2023accurate,lam2023learning,chen2023fengwu,chen2023fuxi,han2024fengwu,zhao2025transforming,zhao2024weathergfm,xu2024generalizing}. 
% rely on reanalysis data
FourcastNet was the first to propose using deep neural networks to learn global atmospheric dynamics~\cite{pathak2022fourcastnet}. 
By scaling up the training stage, Pangu-Weather~\cite{bi2023accurate} and GraphCast~\cite{lam2023learning} simultaneously achieved accuracy levels comparable to those obtained by the operational IFS systems at ECMWF. 
% FuXi~\cite{chen2023fuxi} and FengWu~\cite{chen2023fengwu,han2024fengwu} have further extended the effective prediction period and resolution through multiple weights or replay buffer finetuning. 
% Additionally, GenCast~\cite{price2025probabilistic} explores the use of diffusion models for probabilistic predictions while NeuralGCM~\cite{kochkov2024neural} introduces a physical kernel for hybrid modeling. 
Other works extend the AIWP from aspects including forecasting skill~\cite{chen2023fuxi,chen2023fengwu}, resolution~\cite{han2024fengwu}, probabilistic modelling~\cite{price2025probabilistic}, and physics informed~\cite{kochkov2024neural}.
Although impressive progress has been made, previous AIWP methods still rely on reanalysis data, which introduces inherent limitations, including temporal lag, limited observation use, and dependency on NWP systems.

\textbf{Direct observation prediction. }
% definition % precipitation nowcasting % global dop % irregular data , roll-out and ineffective spatial temporal modelling
Direct observation prediction holds transformative potential for overcoming the dependency on reanalysis, enabling the forecasting of weather using direct observations. 
Although the concept of direct observation prediction has been widely applied in fields such as precipitation nowcasting, where radar echoes are utilized for short-term forecasting~\cite{ravuri2021skilful,zhang2023skilful,gao2023prediff,gong2024postcast,gong2024cascast}, 
its application in global weather prediction remains limited.
In contrast to gridded radar observations, direct observations of the global atmosphere are irregular and non-gridded, making global weather DOP challenging.
% Precipitation nowcasting is a pivot area for direct observation prediction, where regular radar observations are used for nowcasting. 
% In contrast to regular radar observation, direct observations for global atmospheric data are non-gridded and irregular, making global weather DOP challenging.  
Transformer-DOP proposes using a transformer with mask tokens to handle the irregular global Earth observations~\cite{mcnally2024data}. Simultaneously, Graph-DOP employs graph neural networks to flexibly encode direct observations~\cite{alexe2024graphdop}.
EarthNet proposes pretraining the backbone with an observation assimilation task and then finetuning it with a prediction task~\cite{vandal2024global}.  
Although they successfully apply irregular global observation for prediction, these designs suffer from rollout distribution and ineffective spatiotemporal learning as the input space is discrete while the output space is dense.
We propose applying AIDA to transform the input space into a dense one, thereby solving the misalignment between input and output.

% \textbf{Artificial intelligence data assimilation.}
% % ask kun % aida + aiwp % pure aida % aida + aidop
% % data assimilation influenced 
% % observation space -> physical space % 改变模式或者算法 4dvar diffda adas
% % observation space -> observation space earthnet
% The development of data assimilation has also been revolutionized by artificial intelligence. 
% Xiao~\cite{xiao2024towards} was the first to apply the popular traditional numerical data assimilation method, Four-Dimensional Variational, to the AIWP model FengWu.  
% Furthermore, researchers have explored the development of artificial intelligence assimilation methods, such as Fnp~\cite{chen2024fnp} and DiffDA~\cite{huang2024diffda}, which could be applied to both NWP and AIWP models.  
% Although impressive progress has been made, these methods remain limited by reanalysis data and NWP models, which require transforming the observations into physical space.
% Unlike previous methods, EarthNet~\cite{vandal2024global} proposes implementing observation space data assimilation with masked reconstruction. 
% We are motivated to use an observation AIDA model for formulating a completed observation space.  

%% file: Sections/Method.tex
\section{Method}
\label{sec:method}
% 简介DA+WP的思路 
% 整体流程: 压缩 DA WP Mapping + 好处
Our DAWP is designed for global observation forecasting in a uniform satellite observation space. 
The key components of DAWP are an observation space data assimilation module and a cross-regional boundary conditioning weather prediction module.
Additionally, we introduce the mask ViT-VAE to encode observations and produce precipitation variables.  
Our DAWP is illustrated in Figure~\ref{fig:main}.

\subsection{Initialization: observation space assimilation by multi-modal masked autoencoder}
% motivation: 
    % sparse and missing observation 
    %  lead to Disalignment between input and output space -> roll out distribution drift
    % irregular input space -> restricted sptiotmeporal modelling
% We propose using observation space assimilation with a Multi-modal Masked Autoencoder (MMAE)~\cite{bachmann2022multimae,mizrahi20234m} as the initialization stage for direct observation predictions. 
% The missing and sparse observations, attributed to the inherent characteristics of orbital motion, present an irregular input observation space, as shown in the left column of Figure~\ref{fig:main}. 
% Directly taking these satellite observations as inputs not only restricts the network design for spatiotemporal modeling, but also leads to a distribution shift when implementing rollout forecasting, as the output space is required to be a regular one. 
% To meet the gap, we use an MMAE gap-filling areas with blank given contextual information from different sensors and space-time nearby observations.

\begin{figure}[t]
    \centering
    \includegraphics[width=1.0\linewidth]{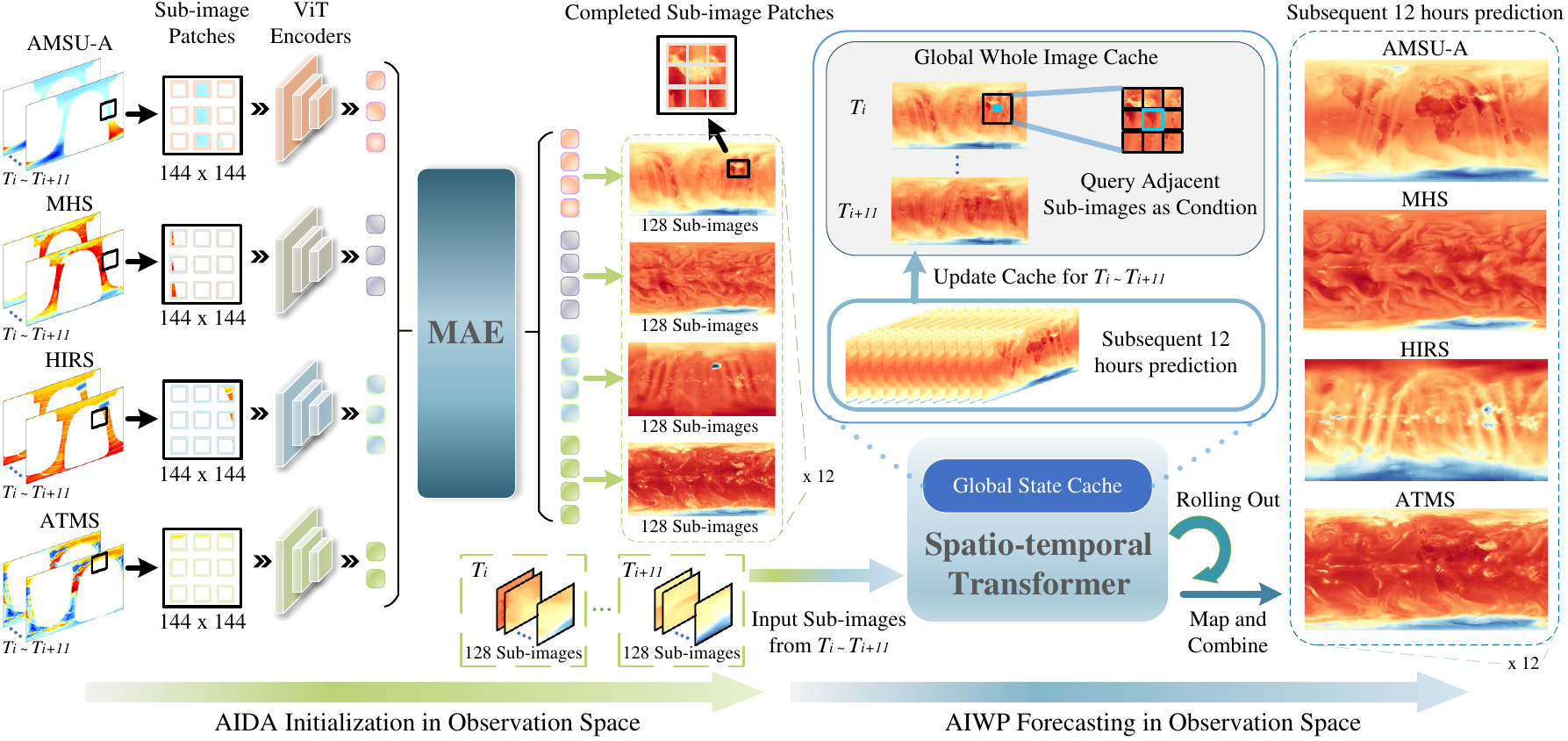}
    \vspace{-0.7cm}
    \caption{The framework of our DAWP. 
    There are two stages in our DAWP: (1) Initialization and (2) Forecasting. 
    % In the initialization stage, an MMAE is used to assimilate sub-images of size 144$\times$144. 
    % The sub-images are tokenized frame by frame with a patch size of 16.
    % Tokens from different satellites within a 12-hour time window are concatenated and passed through the MMAE for assimilation.
    % In the forecasting stage, a sequence of 12 sub-images is input, along with queried adjacent neighbors, for the next 12 hours' prediction.
    % These predicted sub-images are mapped and combined into a global prediction.
    % The global state cache is updated with the global predictions for the next step.
    }
    \label{fig:main}
\end{figure}

The missing and sparse observations, attributed to the inherent characteristics of orbital motion, present an irregular input observation space, as shown in the left column of Figure~\ref{fig:main}. 
Directly taking these satellite observations as inputs not only restricts the network design for spatiotemporal modeling but also leads to a distribution shift when implementing rollout forecasting, as the output space is required to be a regular one. 
To meet the gap, we propose using observation space assimilation with a Multi-modal Masked Autoencoder~\cite{bachmann2022multimae,mizrahi20234m} (MMAE) as the initialization stage for direct observation predictions. 
The MMAE fills in missing areas by leveraging contextual information from different sensors and spatiotemporally nearby observations.
% The observation space with variable missing locations not only restricts the approach for sptiotemporal modeling, but also leads to distribution shift when rolling out as the output space is required to be a regular one. 

% nearby in space-time.
    
% how:
% overview:vmodal-spatial-temporal sequence, training:observation mask, inference: all 

% Our assimilation MMAE are trained across four satellite datasets within random sub regions and a fixed time window. Data from different satellite are first encoded by pretrained satellite-specific mask ViT encoders. 
The core of our assimilation module is a naive MAE that imputes masked multi-modal satellite tokens following ~\cite{he2022masked,bachmann2022multimae,vandal2024global}. 
Since imputation mainly relies on space-time nearby observations from multiple satellite observations, our MMAE sub-regionally processes data from multiple sources within a fixed time window of 12 and a sub-image of 144$\times$144. 
Observations in the time window are tokenized frame by frame by pretrained satellite-specific mask ViT encoders, where missing patches are ignored.
Remaining spatiotemporal tokens from each satellite are concatenated and passed through MAE for complete missing information in the observation space.
For MAE training, we randomly mask a given number of tokens from the whole concatenated remaining tokens and reconstruct the left observed but masked tokens. 
As the number of observed tokens from different time windows can vary, we flexibly pad [EOS] tokens to maintain a uniform sequence length for efficient attention computation. 
In the inference stage, masks for MAE are released to utilize as many available observations as possible.

\subsection{Forecasting: cross-regional boundary conditioning direct observation prediction}
% motivation -> WP based on regional DA requires cross-reginonal interaction
After AIDA, we implement an efficient weather prediction module for direct observation prediction through cross-regional boundary conditioning in the imputed observation space, initialized by our assimilation module. 
For efficiently integrating with the assimilation module pretrained by sub-images, our weather prediction module is also applied to sub-images generated by the assimilation module.
Since sub-images only contain local atmospheric states, predictions on sub-images require cross-regional information interaction for continuous spatiotemporal modeling. 
We introduce a global state cache to store observation states for cross-regional boundary conditioning during forecasting.
% As predictions on sub-images require cross-regional information interaction for continual spatiotemporal modeling, we propose a global state cache to store observation state for cross-regional boundary condition in forecasting.

% structure, how: training, inference: query&update
With the global state cache, our weather prediction module achieves efficient cross-regional boundary conditioning observation forecasting by applying spatiotemporal decoupling attention structure.
In the forecasting stage, the assimilated observations simplify the prediction task into a standardized spatiotemporal forecasting problem, which is widely explored~\cite{gao2022earthformer,tang2024predformer,arnab2021vivit}.
We follow the concept of spatiotemporal decoupling in spatiotemporal forecasting~\cite{gao2022earthformer,gong2024cascast}. 
Specifically, our weather prediction module is composed of $N$ Temporal-Spatial (TS) attention blocks~\cite{arnab2021vivit,tang2024predformer}. 
Attention blocks have the advantages of simplicity and scalability, while the TS spatiotemporal decoupling reduces the sequence length of spatiotemporal tokens for efficient computation.
Utilizing the efficiency of TS decoupling, we simply pad tokens from neighbouring areas to the tokens of the prediction region as SD3~\cite{esser2024scaling} and Flux~\cite{flux2024} do. 
In this way, cross-regional boundary information of the border area is passed to the center forecasting region.
To forecast multistep global observations, we maintain a global state cache during the inference phase, ensuring consistent updates for subsequent steps.
Specifically, when forecasting observations of center sub-images, current adjacent sub-images are queried as conditions from the state cache according to the relative coordinates. 
After one step of global prediction has been completed, the global state cache is updated with the subsequent 12 hours prediction to support rollout forecasting.
More details can be found in the Appendix~\ref{sec:global_state_cache}.

% As the assimilation module inpaints the irregular observations, we could simply formalize the forecasting stage as 

\subsection{Encoding and precipitation mapping via mask ViT-VAE}
% motivation -> irregular multichannel data format
In our DAWP, we introduce a mask ViT-VAE both for encoding and mapping multiple-channel satellite observations with missing values. 
The mask ViT-VAE consists of a vision transformer (ViT) encoder/decoder with masks enabling the model to ignore patches without sufficient observations.
The ViT encoder/decoder provides better compression capability, as detailed in Appendix~\ref{sec:vaes_comparison}, compared to SD-VAE for satellite observations, which typically have more channels than natural images. 
Moreover, it explicitly maintains spatial consistency between tokens and pixels by position embeddings and mitigates the influence of missing values through mask attention.
With our mask Vit-VAE, encoding/decoding is pretrained as a reconstruction task, while the precipitation mapping is trained with ATMS inputs and ATMS-precipitation outputs.
% structure mask vit -> token with mask info, prefer multichannel -> specify mappiing setting

% The structure of our framework is shown in figure~\ref{}

% \subsection{Mask-VAE for the compression of irregular observation}
% % how to compress direct observation with irregular grids 

% \subsection{Assimilation in observation space by multimodal mask autoencoder}
% % learning spatiotemporal correlations between different measure systems; transform data from non-regular observation to a uniform observation space. 

% \subsection{regional direct observation prediction with border condition}
% % the reason for regional prediction; border condition for global prediction (neighbour and state cache) 

% \subsection{precipitation mapping}
% % mapping observation to precipitation

%% file: Sections/Experiment.tex
\section{Experiments}
\label{sec:experiments}
% overview
In the experiment part, a comprehensive analysis of our DAWP is presented. 
First, we introduce the composition of our data and training details. 
Based on the observation data, a comparison of our DAWP with other AI-DOP methods is implemented. 
Further, we evaluate the precipitation forecasting skill of these AI-DOP methods by applying a precipitation mapping network.
In addition to evaluating DAWP's capabilities of forecasting capabilities, we also conducted ablation studies to validate the effectiveness of our modular designs.
Finally, the importance of each satellite for DOP is tested by ablating the input modalities of the assimilation module.

\subsection{Experimental Setups}
% \subsection{Data introduction}
\input{Tables/dataset_info}
\textbf{Data.} The top four datasets listed in Table~\ref{tab:dataset} are used for training our DAWP, while ATMS-precipitation is used to train the precipitation mapping with ATMS.
We generate hourly, 0.16$^\circ$ resolution composites by interpolating and reprojecting the raw data as detailed in Appendix~\ref{sec:data_preproc}. Additional information for dataset split and introduction is presented in Appendix~\ref{sec:dataset_introduction}.
% This composite dataset has a training split with data from January of 2012 to June of 2022 for training and a testing split composed of data from May of 2023 to July of 2023. 
% %  A test part of xxx is held out from training for evaluations. 
% Appendix~\ref{sec:dataset_introduction}.

\textbf{Training details.} The training details are presented in Appendix~\ref{sec:training_details}. 
Training our mask ViT-VAE, MMAE AIDA, and TS decoupling AIWP takes about 1 day, 5 days, and 4 days, respectively. 
Besides, our DAWP is compared with the persistence model, using our AIDA module for completion of missing values, as mentioned in~\cite{gao2022earthformer}. 
We replicate the EarthNet~\cite{vandal2024global} and Transformer-DOP~\cite{ravuri2021skilful} models on our composite dataset, as their codes and data are closed source.

\subsection{Direct observation prediction}
% a time range table 
% line charts for detailed roll out information (done)
\input{Tables/DOP_table}
\begin{figure}[t]
    \centering
    \includegraphics[width=1.0\linewidth]{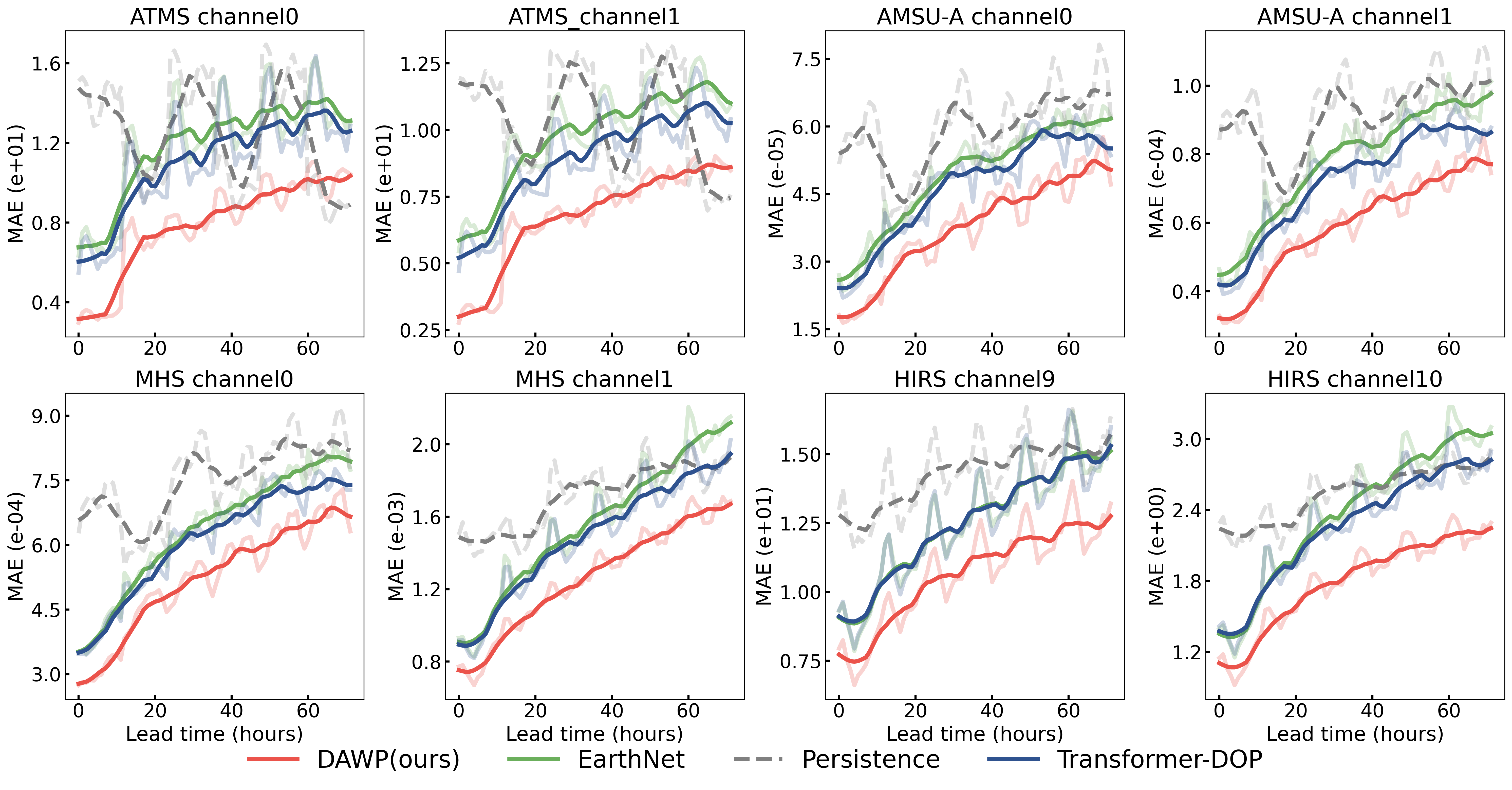}
    \vspace{-0.7cm}
    \caption{Curves of MAE for the prediction of different modalities. The max leadtime is 72h with a 1h temporal resolution.}
    \label{fig:dop}
\end{figure}

\begin{figure}[t]
    \centering
    \includegraphics[width=1.0\linewidth]{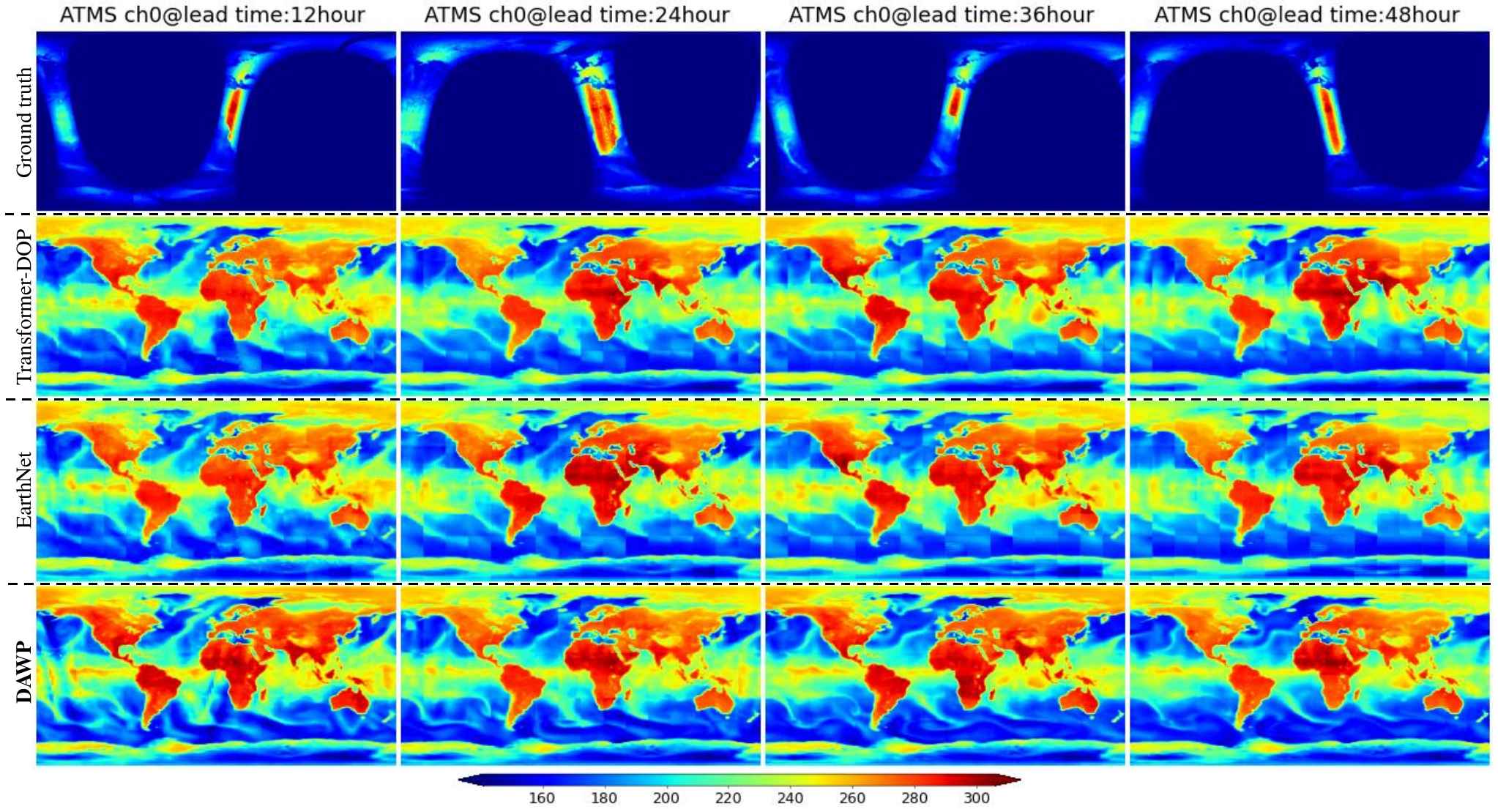}
    \vspace{-0.7cm}
    \caption{A visualization of rollout predictions for global satellite observation forecasting.}
    \label{fig:dop_visualization}
\end{figure}

% dop evaluation setting
We compare our DAWP with other AI-DOP methods to evaluate the ability to predict direct observations.
The metric for evaluation is the mean absolute error (MAE) between the predictions and observed values.
% It is worth noting that the MAE is calculated with the raw observation point by point to ignore the influence of missing values.
The baselines are also trained with sub-images as our DAWP. 
Among them, persistence is a naive method that uses the last observation as the prediction of the next one.
% We apply our assimilation module to construct persistence with completed observations for evaluations on observations with variable missing values.
% EarthNet is trained for prediction with a pretrain stage following~\cite{vandal2024global}. 
% With this setting, we could fairly compare our DAWP with other methods for the direct observation prediction task.

% table analysis
%% 全时间段，全channel，全satellite的领先  %% 24-36h的预测领先其他方法12-24h的预测
The time-averaged MAEs within fixed time windows of 0-12h, 12-24h, and 24-36h are presented in Table~\ref {tab:dop}. 
We select two channels with meaningful patterns of each modality observation to calculate the MAE. 
% The 9th and 10th channels of HIRS are selected, while other modalities' first two channels' MAEs are shown.
The results show that our DAWP significantly outperforms other methods. 
% in all time windows and channels on each modality.
Specifically, our DAPW's MAE on AMSU-A among 24-36h is 3.66 and 5.80, which not only outperforms those of EarthNet and Transformer-DOP, but also surpasses EarthNet's (3.84 and 6.14) and Transformer-DOP's (4.12 and 6.65) MAEs on AMSU-A among 12-24h. 
It is the same when comparing our DAWP with baselines on other modalities.
These results indicate our DAWP has a 12-hour lead time advantage in direct observation prediction.
% figure analysis
%% 介绍图的时间长度时间粒度 %% 周期性的特征  %% 我们的优势从一开始就建立并有随时间扩大的趋势 %% persistent的交点
To provide a more detailed analysis of the observation prediction, we present the MAE figure in a time range of 0-72h in Figure~\ref {fig:dop}. 
It can be observed that the 1h temporal resolution of the figure depicts the periodicity of the prediction errors. 
% As shown in Figure~\ref{fig:dop}, the MAE curve demonstrates oscillatory behavior superimposed on a slowly increasing trend. 
% In subfigure (h), we can observe about 12h intervals between adjacent error peaks. 
This periodicity is attributed to the strip-scanning characteristics of polar-orbiting satellites, which generate periodic observations.
Although there is periodicity in the MAE, our DAWP consistently outperforms other methods, establishing a lead time advantage from the beginning. 
Specifically, in subfigure (a), at the beginning of the prediction, our DAWP's MAE is about 3.5 while the MAE of EarthNet is about 6.8, which is almost 2 times larger than ours.
In addition, EarthNet is surpassed by the naive persistent baseline at the lead time of 16h, but our DAWP doesn't meet the persistent baseline until about 64h.  

% 介绍图的可视化结果
In Figure~\ref{fig:dop_visualization}, we visualize the prediction results of these methods at different lead times. 
At the lead time of 12 hours (the first prediction step), our DAWP exhibits a slightly better prediction than others.
When the lead time increases, a significant distortion appears in the predictions of EarthNet and Transformer-DOP, while our DAWP maintains a relatively stable structure.
It has the same trend as the MAE curve in Figure~\ref{fig:dop}, indicating that our DAWP is more robust in rollout predictions.

% conclusion
%% 直接观测预测的结果表明我们的方法0-72h时间段和所选通道上都优于其他方法，展现了我们的方法在初始预测和rollout预测上的有效性。
The results of direct observation prediction indicate that our DAWP outperforms other methods in the 0-72h time range and selected channels, demonstrating the effectiveness of our method in both initial prediction and rollout prediction. More evaluations on other channels are shown in the Appendix~\ref{sec:DOP_results}.

\subsection{Global precipitation forecasts}
\input{Tables/precipitation_table}
% precipitation evaluation setting 
% In this section, we evaluate the potential of AI-DOP methods for global precipitation forecasting without the requirement of satellite precipitation products.
% To obtain atmospheric variables related to precipitation, we train a precipitation mapping network for transforming direct observation into precipitation products. It is worth noting that in this way, our DAWP does not require satellite precipitation products as input, making the forecasting a quick response to observations.

In this section, we evaluate the potential of AI-DOP methods for global precipitation forecasting. A precipitation mapping network is trained to transform observation predictions into precipitation products. It is worth noting that in this way, our DAWP does not require satellite precipitation products as input, making the forecasting a quick response to observations.

The precipitation skill is evaluated by the Critical Success Index (CSI)~\cite{schaefer1990critical} and False Alarm Ratio (FAR)~\cite{barnes2009corrigendum} metrics on observed points in a time range of 0-12h. 
CSI measures the ability of the model to correctly identify precipitation events, while FAR assesses the reliability of the model's predictions. 
The combined application of CSI and FAR enables a comprehensive precipitation forecasting skill assessment. 
We present a detailed definition of CSI and FAR in Appendix~\ref{sec:metrics_defination}.

% They are core binary classification evaluation metrics that quantify the detection accuracy and reliability of precipitation events.
% CSI is defined as the ratio of the number of true positive (TP) events to the total number of observed events. 
% It measures the ability of the model to correctly identify precipitation events.
% FAR is evaluated to assess the reliability of the model's predictions by calculating the ratio of the number of false positive (FP) events to the total number of predicted events.
% The combined application of CSI and FAR enables a comprehensive precipitation forecasting skill assessment.
% The detailed definition of CSI and FAR is presented in Appendix~\ref{sec:metrics_defination}.

% table analysis
Table~\ref{tab:precipitation} presents the quantitative results of the precipitation forecasting skill of AI-DOP methods on total column water vapor index (TCWV) and surface precipitation (SP).
The thresholds for TCWV and SP are set to [10mm, 20mm, 30mm] and [0.5mm/h, 1.0mm/h, 2.0mm/h], respectively.
%% TCWV csi轻微优势，在所有threshold保持 %% TCWV far明显更低，表明我们的方法误报率低更可靠
On variable TCWV, our DAWP achieves a slight advantage in CSI over other methods. 
Specifically, our DAWP's achieves 0.807 on CSI-30, while Earthnet and transformer-dop achieve 0.786 and 0.789, respectively.
This advantage is maintained across all thresholds.
In terms of FAR, our DAWP is significantly lower than other methods. 
% The FAR-30 of our DAWP is only 0.120, which is roughly 30\% lower than that of Earthnet (0.172) and transformer-dop (0.169).
It indicates that our DAWP predicts more accurate TCWV with lower false alarm rates than baseline methods, demonstrating the reliability of our DAWP.
%% SP csi明显的优势->更准确的预报 
We also analyze the forecasting skill on SP variable. 
The CSI of our DAWP is significantly higher than that of other methods on CSI-0.5 and CSI-1.0, achieving 0.197 and 0.102.
Compared with transformer-dop, our DAWP increases CSI-0.5 and CSI-1.0 by 44.8\% and 78.9\%. 
%% SP CSI 随着阈值增加逐渐减小，earthnet和transformer-dop都已经比persistent低了，丧失预报能力。原因是预测分布便宜到强度更弱的观测空间，以及由此导致的mapping网络的输入分布变化。
Another observation about CSI is a decreasing trend with increasing thresholds. 
Especially, when the threshold increases to 2.0, the CSI of EarthNet and transformer-dop are even lower than that of the naive persistent model.
Only our DAWP maintains a slightly higher forecast skill on CSI 2.0.
The decrease of forecasting skill is caused by the temporal decay of prediction intensity~\cite{ravuri2021skilful,gong2024postcast}, which could further hamper the mapping network's performance.  
%% SP FAR 0.5和1.0的阈值上显著的低于其他方法，在2.0的阈值上略高于earthnet和transformer-dop但明显较persistent低。结合CSI-2.0的结果进行分析，earthnet和transformer-dop没办法产生大于2.0mm的预测，因此FAR低。而DAWP能够在有预测的情况下FAR仅略高于其他方法。
When evaluating FAR on threshold 2.0mm/h, our DAWP uncommonly surpasses Earthnet and transformer-dop slightly. 
It could be explained by considering the CSI-2.0 results, as it is difficult for EarthNet and transformer-dop to produce predictions greater than 2.0mm/h, resulting in few false alarms.
Besides, compared to the persistent baseline with a comparable CSI-2.0 skill, our DAWP has a FAR score which 57.8\% lower than persistent's, showing stronger reliability.
%% conclusion
The evaluation of CSI and FAR scores demonstrates the potential of our DAWP for global precipitation forecasting, simultaneously increasing the accuracy and reliability for precipitation forecasting skill of AI-DOP methods.

\subsection{Ablation study}

\begin{wrapfigure}{r}{0.35\columnwidth}
    \centering
    \vspace{-1cm}
    \includegraphics[width=0.35\columnwidth]{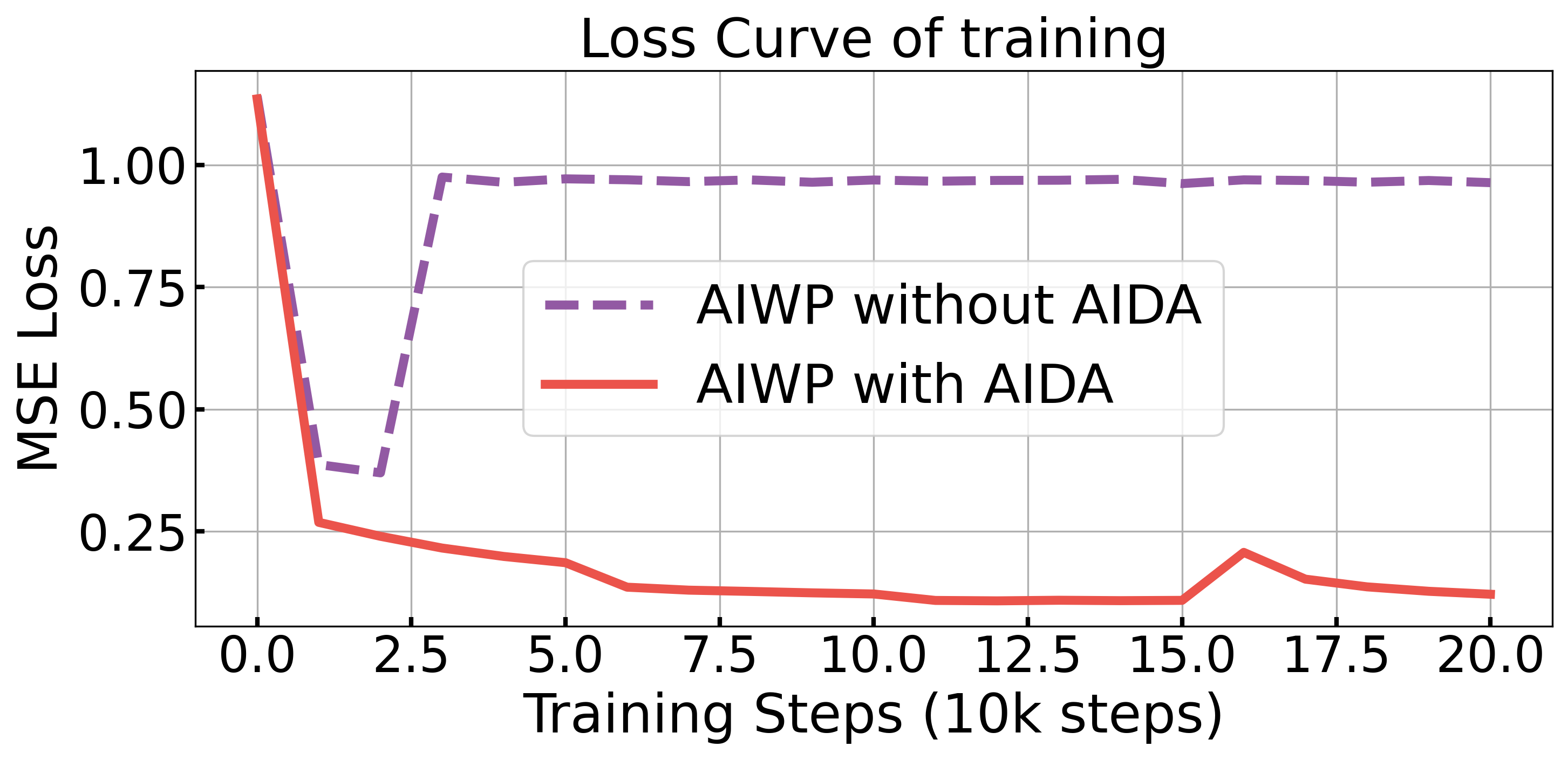} 
    \vspace{-0.7cm}
    \caption{
    % The curve of loss for training our AIWP module directly in an irregular observation space and an initialized observation space constructed by AIDA.
    Training loss curve.
    }
    \label{fig:AIDA_efficient}
    \vspace{-0.5cm}
\end{wrapfigure}

\textbf{Effect of AIDA initialization.} 
An ablation study is conducted to validate the effectiveness of our AIDA module on our DAWP and other AI-DOP methods.
The results demonstrate that AIDA enables efficient spatiotemporal modeling and enhances the rollout prediction ability of AI-DOP methods by imputing the observation space.

\begin{wrapfigure}{r}{0.5\columnwidth}
    \centering
    \vspace{-0.4cm}
    \includegraphics[width=0.5\columnwidth]{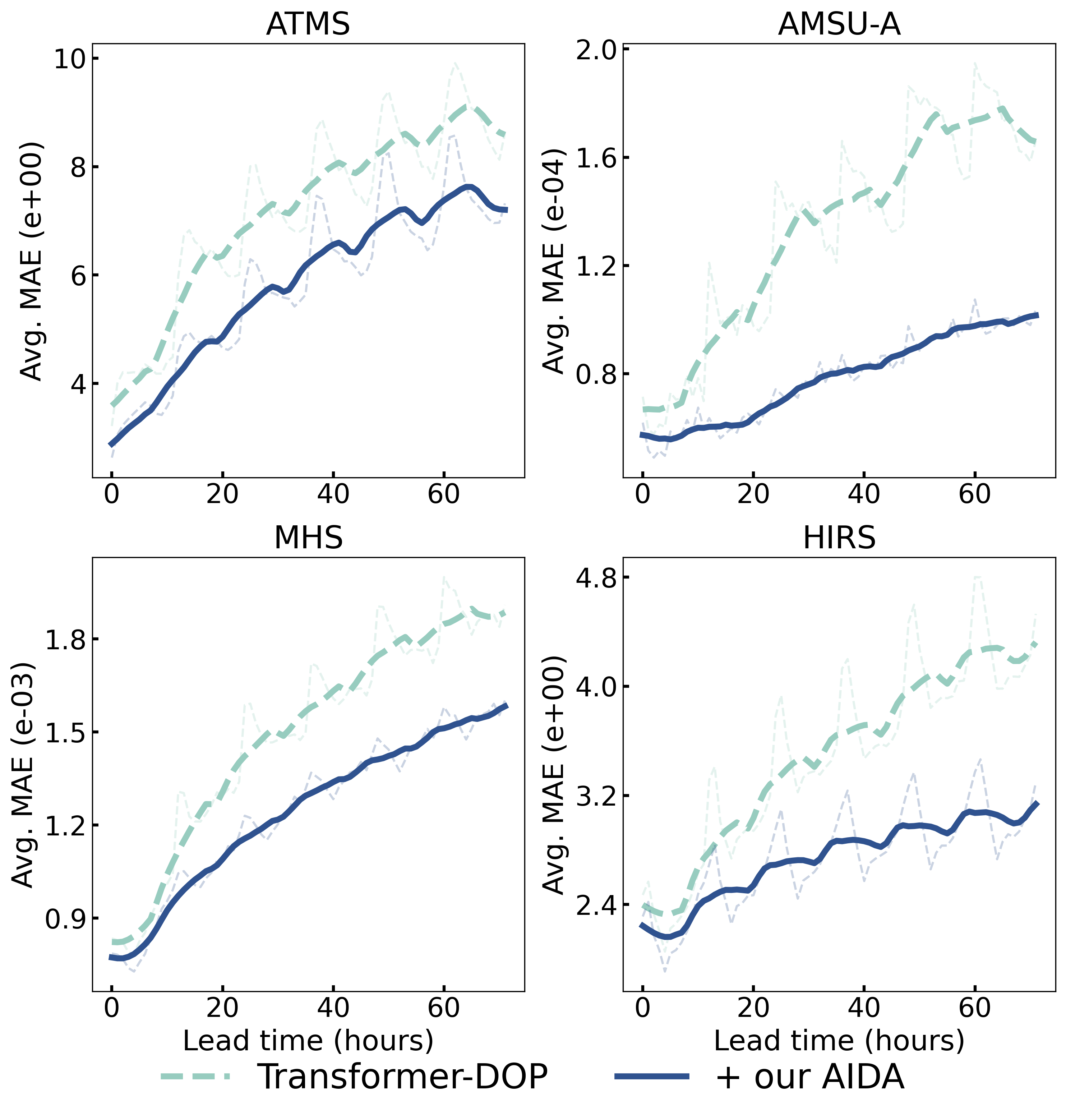}
    \vspace{-0.8cm}
    \caption{
    % The MAE curves of different modalities on Transformer-DOP with(w/o) our AIDA initialization.
    The MAE curves of Transformer-DOP with(w/o) our AIDA initialization.
    }
    \label{fig:AIDA_rollout}
    \vspace{-0.4cm}
\end{wrapfigure}

% setting: cancel the AIDA stage of our DAWP
% phenomenon: unstable training
% reason: classical spatiotemporal learning methods are difficult to learn the dynamics in a observation space with variable missing values.
% conclusion: AIDA初始化规整observation space有助于进行高效的时空学习建模
The training is unstable after cancelling the AIDA stage in our DAWP.
We show the curve of training loss in Figure~\ref{fig:AIDA_efficient}.
It can be observed that the training loss of our DAWP with AIDA is decreasing steadily, while the loss curve without the initialization of AIDA dramatically increases at about 20k training steps and maintains a MSE loss of 1.0.
In contrast, the loss of our DAWP with AIDA could converge to less than 0.1 after 200k training steps.
This phenomenon indicates that the classical spatiotemporal learning methods are difficult to learn the dynamics in an observation space with variable missing values.
By imputing the observation space into a completed spatiotemporal space, AIDA benefits efficient spatiotemporal learning modeling.

% setting: training transformer-dop after AIDA initialization
% phenomenon: rollout prediction is more accurate than the original transformer-dop
% reason: AIDA intilization 缓解了distribution drift caused by the gap between observation with missing values and the completed rollout inputs.
% conclusion: AIDA initialization boosts AI-DOP model for multi-step rollout prediction.
Applying AIDA initialization enhances the rollout prediction ability of AI-DOP methods. 
The quantitative results of transformer-dop are depicted in Figure~\ref {fig:AIDA_rollout}. 
We exhibit the averaged MAE of 0-72h prediction on each sensor with/without AIDA initialization.
Transformer-DOP with AIDA initialization significantly outperforms the original Transformer-DOP when the lead time increases. 
It reveals the potential of AIDA initialization for boosting the AI-DOP model for multi-step rollout prediction.

\textbf{Gains of cross-regional boundary conditioning. }
% overview: 验证cross-regional condition的有效性。更准确的region预测结果，区域间更连续的预测效果。实验设置是对比有无cross-regional condition的结果。
We compare the observation predictions with and without cross-regional conditions by training a DAWP without neighbour regions as inputs.
This ablation study is conducted to verify the effect on the accuracy and spatial continuity of the prediction.

\begin{wrapfigure}{r}{0.5\columnwidth}
    \centering
    \vspace{-0.5cm}
    \includegraphics[width=0.5\columnwidth]{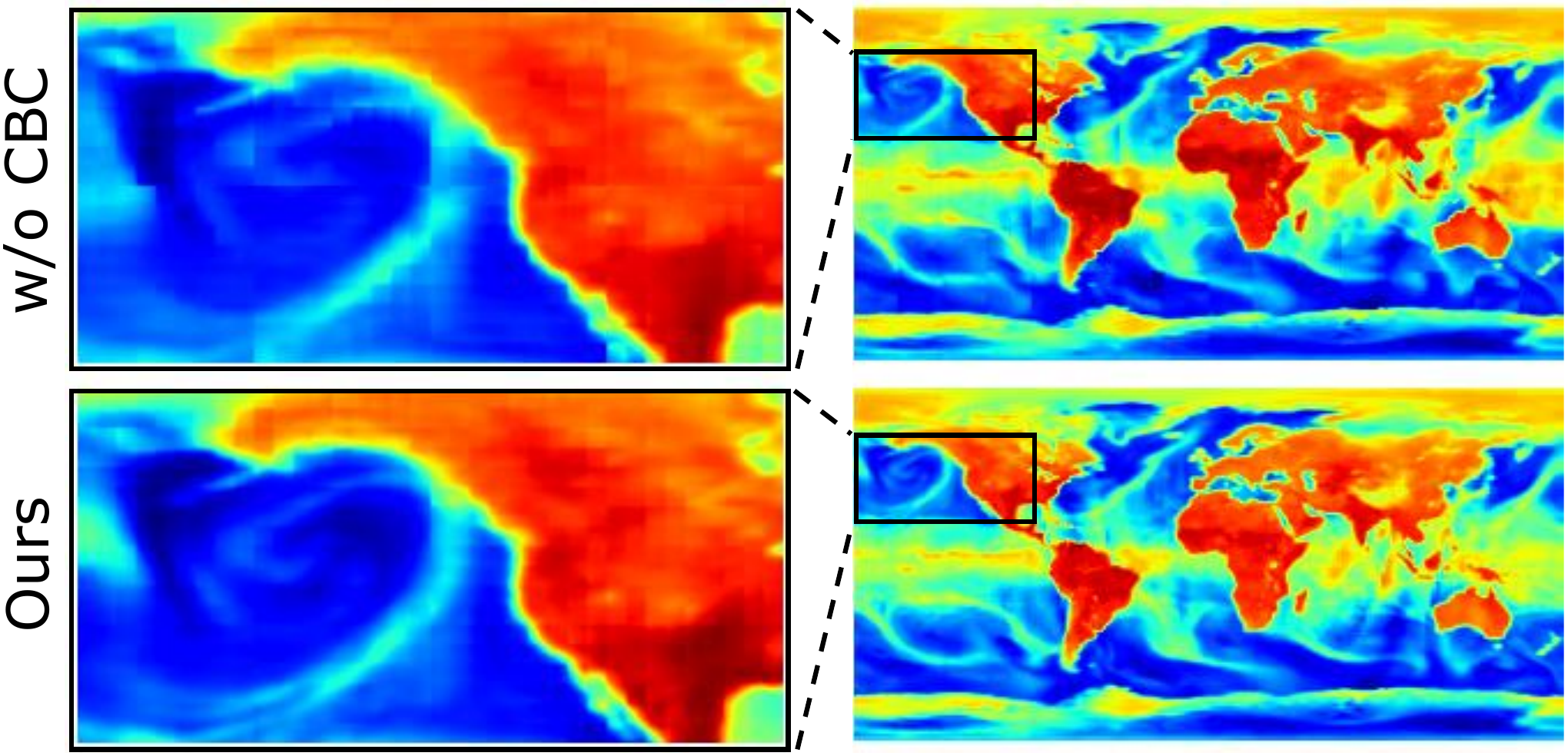}
    \vspace{-0.7cm}
    \caption{A visualization of forecasting results with(w/o) Cross-regional Boundary Conditioning.}
    \label{fig:CBC}
    \vspace{-0.2cm}
\end{wrapfigure}

% 描述表
The convergence loss of our DAWP with and without cross-regional condition is presented in Table~\ref{tab:CBC}, showing conditioning on neighbour regions improves the prediction accuracy.
In this table, we evaluate the convergence loss of center areas and neighbor areas. 
For DAWP without Cross-regional Boundary Conditioning (CBC), we take the center area as the neighbour area.
% w center > w region, w center > wo center
The average convergence loss of the center area is 0.106, which is significantly than that of the neighbour area.
Besides, it also significantly outperforms the DAWP without CBC on the centre area by about 15.6\%. 
% 原因是由于大气的物理运动边界信息对于weather prediction很重要
This result validates the necessity of cross-regional conditions for weather prediction, as the boundary information of atmospheric physical motion is crucial for accurate weather forecasting.
% 说明了cross-regional 对预测准确性的提升

% 描述图
Another benefit of cross-regional conditions is the improvement of prediction continuity, as shown in Figure~\ref{fig:CBC}.
The first and second rows show the predictions of ATMS channel 0 at a lead time of 6 hours without and with CBC, respectively.
It can be observed in the black box that the continuity of the adjacent regions is improved, which is helpful for keeping the atmospheric structure. 

\input{Tables/boarder_aware_table}

% \subsubsection{Compare mask ViT-VAE with SD-VAE}
% \input{Tables/vae_table}
% % setting: 对比SD-VAE在压缩多channel有缺失观测的satellite数据的能力 
% % 介绍表
% % multichannel 上优秀, 少channel时compare
% % mask 对稀疏观测的影响
% We explore the ability of our mask ViT-VAE to compress satellite data with multiple channels and missing values by comparisons with other VAEs.

% The results are shown in Table~\ref{tab:vae}, where we compare reconstruction loss of different VAEs.
% First, VAEs with ViT structure is more effective for reconstructing modality data with multiple channels such as HIRS and AMSU-A, 
% while on modalities with fewer channels, there are only slight increase of reconstruction loss.
% Another observation is that the application of mask consistently increase VAEs' reconstruction ability on HIRS.
% For ViT-VAE, it is beneficial to use mask for the compuation of attention between patch tokens, 
% as it could directly weaken the influence of missing tokens.  

% drop one inference 
% one to many inference
% \input{Tables/drop_one}
% \input{Tables/one2many}

\begin{figure}[t]
    \centering
    \includegraphics[width=1.0\linewidth]{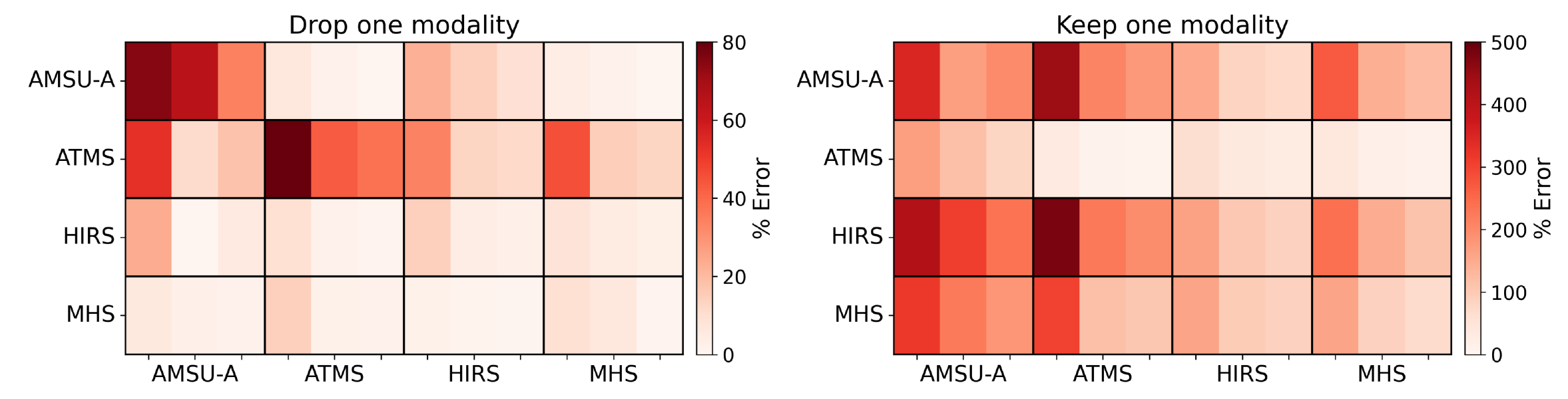}
    \vspace{-0.7cm}
    \caption{Matrix of relative errors under the setting of dropping one modality and keeping one modality. 
    The three columns in a black rectangle represent the relative MAE error ratios of 0-12h, 12-24h, and 24-36h lead times, respectively.}
    \label{fig:exp_sensitivity}
\end{figure}

% 目的: gain insight of the importance for each satellite's data in observation prediction
% seeting: keep one and drop one
%% drop one: 在AIDA之前去掉一个satellite的观测数据进行多步推测
%% keep one: 在AIDA之前只保留一个satellite的观测数据进行多步推测
%% 扣insight
\textbf{Modality sensitivity analysis.} 
We introduce the experiment of using different modality combinations for multistep observation prediction to gain insight into the importance of each modality's data.
As shown in Figure~\ref{fig:exp_sensitivity}, a drop one and a keep one combination are explored.
The setting of drop one means removing one modality's observation before AIDA initialization and conducting multi-step prediction, while in the keep one setting,
we only keep one modality's observation before AIDA initialization for forecasting.
We evaluate the influence of each modality by calculating the relative MAE error ratios between the MAE of DAWP with completed modality inputs. 

% 介绍drop one的实验结果 -> 结论
%% 介绍drop one figure的row column含义，每个小块代表一个时间窗口
%% 总体趋势: 衰减的误差->长期预测使用的信息还不足够，变差的MAE->依赖所有
%% HIRS与MHS能够在没有自身的时候预测得比较好，信息冗余
The result of drop one is shown in the left of Figure~\ref{fig:exp_sensitivity}. 
Row names of the figure indicate the modality that is dropped, 
while columns sequentially included in a modality name represent the relative MAE error ratio of this modality in time windows 0-12h, 12-24h, and 24-36h.
The overall trend of the drop one setting demonstrates that dropping any modality's data leads to an increase of MAE error.
Besides, it's observed that for the rollout predictions, the MAE error ratio is gradually decreasing, indicating the insufficient usage of observations for multistep predictions.
When focusing on single modalities, we find that HIRS and MHS can still achieve a relatively low MAE error ratio when their own data is dropped, indicating that they have redundant information for the prediction.

% 介绍keep one的实验结果 -> 结论
%% figure 含义同dropone
%% ATMSerror ratio最低ATMS信息较为完备: a substantial amount of information for predicting all modalities
%% AMSU-A自预测能力最弱，动力信息不足
The error ratio matrix of keep one setting is also shown in Figure~\ref{fig:exp_sensitivity}.
When only keeping ATMS observation, other modalities' MAE error ratios of predictions are the lowest, which are even lower than keeping the satellites themselves.
This indicates that ATMS has a substantial amount of information for prediction.
In contrast, for predicting one modality itself, AMSU-A exhibits the highest MAE error ratio, showing that it has the least spatiotemporal dynamics information.

%% file: Tables/dataset_info.tex
\begin{table}[t]
\caption{Dataset overview.}
\resizebox{\linewidth}{!}{
\begin{tabular}{lcccc}
\toprule
Modality/Sensor & Satellite & Channels & Level & Period \\ \midrule
Advanced Microwave Sounding Unit-A (AMSU-A)~\cite{eumetsat}  & NOAA18 \& 19 & Microwave radiance 15 bands & 1B & 2007-2023 \\
Advanced Technology Microwave Sounder (ATMS)~\cite{npp_atms,noaa_atms} & NPP \& NOAA20 & Brightness temperature 9 bands & 1C & 2012-2023 \\
High Resolution Infrared Radiation Sounder (HIRS)~\cite{eumetsat} & NOAA18 \& 19 & Infrared radiance 20 bands & 1B & 2007-2023 \\ 
Microwave Humidity Sounder (MHS)~\cite{eumetsat} & NOAA18 \& 19 & Microwave radiance 5 bands & 1B & 2007-2023 \\ \midrule
ATMS-Precipitation~\cite{npp_precipitation,noaa_precipitation} & NPP \& NOAA20 & Precipitation product 2 channels & 2A & 2012-2023 \\ \bottomrule
\end{tabular}
}
\label{tab:dataset}
\end{table}

%% file: Tables/DOP_table.tex
% \usepackage{multirow}
\begin{table}[t]\footnotesize
\centering
\caption{MAE error of forecasting during 3 lead time periods (0-12h, 12-24h, and 24-36h) for different channels of the satellite data. 
We use the unit of 1e-5 for AMSU-A, 1e-4 for MHS, and 1e-0 for both ATMS and HIRS.}
\begin{tabular}{l|c|cc|cc|cc|cc}
\toprule
                             &                             & \multicolumn{2}{c|}{AMSU-A}                                 & \multicolumn{2}{c|}{ATMS}                                     & \multicolumn{2}{c|}{HIRS}                                    & \multicolumn{2}{c}{MHS}                                      \\ \cline{3-10} 
\multirow{-2}{*}{Methods}    & \multirow{-2}{*}{Lead time} & \cellcolor[HTML]{EFEFEF}ch0  & \cellcolor[HTML]{DAD7D7}ch1  & \cellcolor[HTML]{EFEFEF}ch0   & \cellcolor[HTML]{DAD7D7}ch1   & \cellcolor[HTML]{EFEFEF}ch9   & \cellcolor[HTML]{DAD7D7}ch10  & \cellcolor[HTML]{EFEFEF}ch0  & \cellcolor[HTML]{DAD7D7}ch1   \\ \midrule
Persistence~\cite{gao2022earthformer}                  &                             & \cellcolor[HTML]{EFEFEF}5.86 & \cellcolor[HTML]{DAD7D7}9.15 & \cellcolor[HTML]{EFEFEF}14.37 & \cellcolor[HTML]{DAD7D7}11.69 & \cellcolor[HTML]{EFEFEF}12.43 & \cellcolor[HTML]{DAD7D7}2.21 & \cellcolor[HTML]{EFEFEF}7.01 & \cellcolor[HTML]{DAD7D7}14.74 \\
EarthNet~\cite{vandal2024global}                     &                             & \cellcolor[HTML]{EFEFEF}2.93 & \cellcolor[HTML]{DAD7D7}4.89 & \cellcolor[HTML]{EFEFEF}6.96  & \cellcolor[HTML]{DAD7D7}6.14  & \cellcolor[HTML]{EFEFEF}9.15  & \cellcolor[HTML]{DAD7D7}1.39 & \cellcolor[HTML]{EFEFEF}3.96 & \cellcolor[HTML]{DAD7D7}9.65  \\
Transformer-DOP~\cite{mcnally2024data}              &                             & \cellcolor[HTML]{EFEFEF}2.67 & \cellcolor[HTML]{DAD7D7}4.48 & \cellcolor[HTML]{EFEFEF}6.40   & \cellcolor[HTML]{DAD7D7}5.61  & \cellcolor[HTML]{EFEFEF}9.22  & \cellcolor[HTML]{DAD7D7}1.42 & \cellcolor[HTML]{EFEFEF}3.91 & \cellcolor[HTML]{DAD7D7}9.48  \\
\cellcolor[HTML]{CBCEFB}Ours & \multirow{-4}{*}{0-12h}     & \cellcolor[HTML]{CBCEFB}1.92 & \cellcolor[HTML]{CBCEFB}3.39 & \cellcolor[HTML]{CBCEFB}3.36  & \cellcolor[HTML]{CBCEFB}3.27  & \cellcolor[HTML]{CBCEFB}7.70   & \cellcolor[HTML]{CBCEFB}1.12 & \cellcolor[HTML]{CBCEFB}3.07 & \cellcolor[HTML]{CBCEFB}7.91  \\ \midrule
Persistence~\cite{gao2022earthformer}                 &                             & \cellcolor[HTML]{EFEFEF}4.35 & \cellcolor[HTML]{DAD7D7}6.94 & \cellcolor[HTML]{EFEFEF}10.40  & \cellcolor[HTML]{DAD7D7}8.86  & \cellcolor[HTML]{EFEFEF}13.58 & \cellcolor[HTML]{DAD7D7}2.30  & \cellcolor[HTML]{EFEFEF}6.07 & \cellcolor[HTML]{DAD7D7}15.09 \\
EarthNet~\cite{vandal2024global}                      &                             & \cellcolor[HTML]{EFEFEF}4.12 & \cellcolor[HTML]{DAD7D7}6.65 & \cellcolor[HTML]{EFEFEF}11.25 & \cellcolor[HTML]{DAD7D7}9.00  & \cellcolor[HTML]{EFEFEF}11.14 & \cellcolor[HTML]{DAD7D7}1.98 & \cellcolor[HTML]{EFEFEF}5.46 & \cellcolor[HTML]{DAD7D7}13.11 \\
Transformer-DOP~\cite{mcnally2024data}              &                             & \cellcolor[HTML]{EFEFEF}3.84 & \cellcolor[HTML]{DAD7D7}6.14 & \cellcolor[HTML]{EFEFEF}10.04 & \cellcolor[HTML]{DAD7D7}8.04  & \cellcolor[HTML]{EFEFEF}11.08 & \cellcolor[HTML]{DAD7D7}1.95 & \cellcolor[HTML]{EFEFEF}5.19 & \cellcolor[HTML]{DAD7D7}12.65 \\
\cellcolor[HTML]{CBCEFB}Ours & \multirow{-4}{*}{12-24h}    & \cellcolor[HTML]{CBCEFB}3.11 & \cellcolor[HTML]{CBCEFB}5.12 & \cellcolor[HTML]{CBCEFB}7.35  & \cellcolor[HTML]{CBCEFB}6.35  & \cellcolor[HTML]{CBCEFB}9.57  & \cellcolor[HTML]{CBCEFB}1.54 & \cellcolor[HTML]{CBCEFB}4.51 & \cellcolor[HTML]{CBCEFB}10.54 \\ \midrule
Persistence~\cite{gao2022earthformer}                 &                             & \cellcolor[HTML]{EFEFEF}6.39 & \cellcolor[HTML]{DAD7D7}9.84 & \cellcolor[HTML]{EFEFEF}15.37 & \cellcolor[HTML]{DAD7D7}12.52 & \cellcolor[HTML]{EFEFEF}14.61 & \cellcolor[HTML]{DAD7D7}2.61 & \cellcolor[HTML]{EFEFEF}8.00 & \cellcolor[HTML]{DAD7D7}17.86 \\
EarthNet~\cite{vandal2024global}                     &                             & \cellcolor[HTML]{EFEFEF}5.17 & \cellcolor[HTML]{DAD7D7}8.14 & \cellcolor[HTML]{EFEFEF}12.52 & \cellcolor[HTML]{DAD7D7}10.08 & \cellcolor[HTML]{EFEFEF}12.37 & \cellcolor[HTML]{DAD7D7}2.36 & \cellcolor[HTML]{EFEFEF}6.41 & \cellcolor[HTML]{DAD7D7}15.08 \\
Transformer-DOP~\cite{mcnally2024data}              &                             & \cellcolor[HTML]{EFEFEF}4.91 & \cellcolor[HTML]{DAD7D7}7.54 & \cellcolor[HTML]{EFEFEF}11.35 & \cellcolor[HTML]{DAD7D7}9.07  & \cellcolor[HTML]{EFEFEF}12.39 & \cellcolor[HTML]{DAD7D7}2.27 & \cellcolor[HTML]{EFEFEF}6.22 & \cellcolor[HTML]{DAD7D7}14.70  \\
\cellcolor[HTML]{CBCEFB}Ours & \multirow{-4}{*}{24-36h}    & \cellcolor[HTML]{CBCEFB}3.66 & \cellcolor[HTML]{CBCEFB}5.80  & \cellcolor[HTML]{CBCEFB}7.84  & \cellcolor[HTML]{CBCEFB}6.81  & \cellcolor[HTML]{CBCEFB}10.71 & \cellcolor[HTML]{CBCEFB}1.79 & \cellcolor[HTML]{CBCEFB}5.15 & \cellcolor[HTML]{CBCEFB}12.22 \\ \bottomrule
\end{tabular}
\label{tab:dop}
\end{table}

%% file: Tables/precipitation_table.tex
\begin{table}[t]
\centering
\caption{Forecasting skills in 12 hours on precipitation-related variables Total Column Water Vapor (TCWV) and Surface Precipitation (SP).
CSI and FAR scores are calculated on different thresholds.}
\resizebox{\linewidth}{!}{
\begin{tabular}{l|ccc|ccc|ccc|ccc}
\toprule
                         & \multicolumn{6}{c|}{TCWV (mm)}                       & \multicolumn{6}{c}{SP (mm/h)}                             \\ 
\multirow{-2}{*}{Method}                        & CSI-10 & CSI-20 & CSI-30 & FAR-10 & FAR-20 & FAR-30 & CSI-0.5 & CSI-1.0 & CSI-2.0 & FAR-0.5 & FAR-1.0 & FAR-2.0 \\ \midrule
Persistence~\cite{gao2022earthformer}             & 0.853  & 0.702  & 0.636  & 0.121  & 0.266  & 0.332  & 0.110    & 0.073   & 0.031   & 0.844   & 0.861   & 0.655   \\
EarthNet~\cite{vandal2024global}                & 0.909  & 0.822  & 0.786  & 0.047  & 0.130   & 0.172  & 0.127   & 0.050    & 0.008   & 0.666   & 0.692   & \textbf{0.180}    \\
Transformer-DOP~\cite{mcnally2024data}         & 0.905  & 0.822  & 0.789  & 0.053  & 0.130   & 0.169  & 0.136   & 0.057   & 0.010    & 0.655   & 0.685   & 0.214   \\
Ours                    & \textbf{0.917}  & \textbf{0.841}  & \textbf{0.807}  & \textbf{0.034}  & \textbf{0.088}  & \textbf{0.120}   & \textbf{0.197}   & \textbf{0.102}   & \textbf{0.035}   & \textbf{0.529}   & \textbf{0.577}   & 0.283  \\
\bottomrule
\end{tabular}
}
\label{tab:precipitation}
\end{table}

%% file: Tables/boarder_aware_table.tex
\begin{wraptable}{r}{0.5\textwidth}
    \vspace{-0.7cm}
    \centering
    \caption{Converged loss of our DAWP with and without CBC module. The average loss is calculated by averaging the loss of all four modalities. }
    \resizebox{0.5\textwidth}{!}{
    \begin{tabular}{l|c|cccc|c}
    \toprule
                               &                                & \multicolumn{4}{c|}{Modalitiy}                                                                                                &                               \\ %\cline{3-6}
    \multirow{-2}{*}{Module}   & \multirow{-2}{*}{Area}         & AMSU-A                        & ATMS                          & HIRS                          & MHS                           & \multirow{-2}{*}{Avg.}        \\ \hline
                                & border & 0.063 & 0.101 & 0.236 & 0.111 & 0.128 \\ 
    \multirow{-2}{*}{w/o CBC}  & center & 0.063 & 0.101 & 0.236 & 0.111 & 0.128 \\ \hline
                                & border & 0.069 & 0.098 & 0.324 & 0.101 & 0.148 \\
    \multirow{-2}{*}{with CBC}  & center & \textbf{0.054} &\textbf{0.074} & \textbf{0.214} & \textbf{0.084} & \textbf{0.106} \\ \bottomrule
    \end{tabular}
    }
    \label{tab:CBC}
\vspace{-0.5cm}
\end{wraptable}

%% file: Sections/Conclusion.tex
\section{Conclusion}
In this paper, we propose DAWP, a novel framework using AIWP for observation prediction with an AIDA module as initialization. 
Comprehensive experiments are conducted to validate the efficiency and potential of our DAWP framework for observation forecasting and downstream applications such as precipitation forecasting.
\textbf{Broader Impacts\&Future Work}: First, our framework readily integrates variable observations, demonstrating its potential as an implicit Earth system modeling framework. 
Second, our framework has broad application prospects. 
It can seamlessly adapt to diverse downstream tasks-such as surface parameter estimation, wildfire monitoring, and sea ice mapping-whenever observations or retrieval operators are available, similar to precipitation forecasting.
% As long as there are observations or retrieval operators, our framework enables seamless adaptation across surface parameter estimation, wildfire monitoring, and sea ice mapping, similar to precipitation forecasting.
Third, our framework holds a promising potential for directly predicting physical variables by integrating observations of weather variables such as station data. 
% there is potential for our framework to directly predict physical variables by integrating observations of weather variables such as station data. 
\textbf{Limitations}: Although our DAWP framework improves the observation forecasting, the sources of observation are still homogeneous in satellite observations. More observation sources will be integrated with DAWP in the future.

% The AIDA module assimilates different satellite observations with an MMAE.
% The observations with variable missing values are encoded by our mask ViT-VAE as input tokens of the AIDA module.
% In the completed observation space initialized by AIDA, a transformer with spatiotemporal decoupling attentions is used to predict the future observations.
% Additionally, we introduce cross-regional boundary conditioning to improve the global satellite observation prediction based on sub-images.
% The experimental results demonstrate the importance of constructing a complete observation space for efficient observation prediction, which could motivate the development of observation forecasting.
% Although the DAWP framework enhances the observation prediction, the sources of observation are still homogeneous in satellite observations. 
% We believe that with more observations from different devices and locations, there is potential to build skillful observation forecasting systems for weather prediction. 
\label{sec:conclusion}

%% file: checklist.tex
%%%%%%%%%%%%%%%%%%%%%%%%%%%%%%%%%%%%%%%%%%%%%%%%%%%%%%%%%%%%

\newpage
\section*{NeurIPS Paper Checklist}

%%% BEGIN INSTRUCTIONS %%%
The checklist is designed to encourage best practices for responsible machine learning research, addressing issues of reproducibility, transparency, research ethics, and societal impact. Do not remove the checklist: {\bf The papers not including the checklist will be desk rejected.} The checklist should follow the references and follow the (optional) supplemental material.  The checklist does NOT count towards the page
limit. 

Please read the checklist guidelines carefully for information on how to answer these questions. For each question in the checklist:
\begin{itemize}
    \item You should answer \answerYes{}, \answerNo{}, or \answerNA{}.
    \item \answerNA{} means either that the question is Not Applicable for that particular paper or the relevant information is Not Available.
    \item Please provide a short (1–2 sentence) justification right after your answer (even for NA). 
   % \item {\bf The papers not including the checklist will be desk rejected.}
\end{itemize}

{\bf The checklist answers are an integral part of your paper submission.} They are visible to the reviewers, area chairs, senior area chairs, and ethics reviewers. You will be asked to also include it (after eventual revisions) with the final version of your paper, and its final version will be published with the paper.

The reviewers of your paper will be asked to use the checklist as one of the factors in their evaluation. While "\answerYes{}" is generally preferable to "\answerNo{}", it is perfectly acceptable to answer "\answerNo{}" provided a proper justification is given (e.g., "error bars are not reported because it would be too computationally expensive" or "we were unable to find the license for the dataset we used"). In general, answering "\answerNo{}" or "\answerNA{}" is not grounds for rejection. While the questions are phrased in a binary way, we acknowledge that the true answer is often more nuanced, so please just use your best judgment and write a justification to elaborate. All supporting evidence can appear either in the main paper or the supplemental material, provided in appendix. If you answer \answerYes{} to a question, in the justification please point to the section(s) where related material for the question can be found.

IMPORTANT, please:
\begin{itemize}
    \item {\bf Delete this instruction block, but keep the section heading ``NeurIPS Paper Checklist"},
    \item  {\bf Keep the checklist subsection headings, questions/answers and guidelines below.}
    \item {\bf Do not modify the questions and only use the provided macros for your answers}.
\end{itemize}

%%% END INSTRUCTIONS %%%

\begin{enumerate}

\item {\bf Claims}
    \item[] Question: Do the main claims made in the abstract and introduction accurately reflect the paper's contributions and scope?
    \item[] Answer: \answerYes{}% Replace by \answerYes{}, \answerNo{}, or \answerNA{}.
    \item[] Justification: %\justificationTODO{}
    We clarify our contributions in the abstract and introduction.

\item {\bf Limitations}
    \item[] Question: Does the paper discuss the limitations of the work performed by the authors?
    \item[] Answer: \answerYes{} % Replace by , \answerNo{}, or \answerNA{}.
    \item[] Justification: %\justificationTODO{}
    We discuss the limitations of our work in the conclusion section.

\item {\bf Theory assumptions and proofs}
    \item[] Question: For each theoretical result, does the paper provide the full set of assumptions and a complete (and correct) proof?
    \item[] Answer: \answerNA{}, % Replace by \answerYes{}, \answerNo{}, or \answerNA{}.
    \item[] Justification: %\justificationTODO{}
    This is a paper for applications.

    \item {\bf Experimental result reproducibility}
    \item[] Question: Does the paper fully disclose all the information needed to reproduce the main experimental results of the paper to the extent that it affects the main claims and/or conclusions of the paper (regardless of whether the code and data are provided or not)?
    \item[] Answer: \answerYes{} % Replace by \answerYes{}, \answerNo{}, or \answerNA{}.
    \item[] Justification: %\justificationTODO{}
    We provide all experimental settings for reproductivity.

\item {\bf Open access to data and code}
    \item[] Question: Does the paper provide open access to the data and code, with sufficient instructions to faithfully reproduce the main experimental results, as described in supplemental material?
    \item[] Answer: \answerYes{} % Replace by \answerYes{}, \answerNo{}, or \answerNA{}.
    \item[] Justification: %\justificationTODO{}
    We train our model on ATMS, AMSU-A, MHS, and HIRS, which could be obtained publicly on their websites. The detailed settings for data resampling are provided in supplemental material. We will release our code on github.

\item {\bf Experimental setting/details}
    \item[] Question: Does the paper specify all the training and test details (e.g., data splits, hyperparameters, how they were chosen, type of optimizer, etc.) necessary to understand the results?
    \item[] Answer: \answerYes{} % Replace by \answerYes{}, \answerNo{}, or \answerNA{}.
    \item[] Justification: %\justificationTODO{}
    The complete experimental settings, including the dataset, hyperparameters, type of optimizer, etc, are provided in the Experiments section and supplemental materials.

\item {\bf Experiment statistical significance}
    \item[] Question: Does the paper report error bars suitably and correctly defined or other appropriate information about the statistical significance of the experiments?
    \item[] Answer: \answerNo{} % Replace by \answerYes{}, \answerNo{}, or \answerNA{}.
    \item[] Justification: %\justificationTODO{}
    The paper does not report error bars, like other related work.

\item {\bf Experiments compute resources}
    \item[] Question: For each experiment, does the paper provide sufficient information on the computer resources (type of compute workers, memory, time of execution) needed to reproduce the experiments?
    \item[] Answer: \answerYes{} % Replace by \answerYes{}, \answerNo{}, or \answerNA{}.
    \item[] Justification: %\justificationTODO{}
    The sufficient information on the computer resources is provided in the Experimets section
    
    
\item {\bf Code of ethics}
    \item[] Question: Does the research conducted in the paper conform, in every respect, with the NeurIPS Code of Ethics \url{https://neurips.cc/public/EthicsGuidelines}?
    \item[] Answer: \answerYes{} % Replace by \answerYes{}, \answerNo{}, or \answerNA{}.
    \item[] Justification: %\justificationTODO{}
    We perform this work following the NeurIPS Code of Ethics exactly.

\item {\bf Broader impacts}
    \item[] Question: Does the paper discuss both potential positive societal impacts and negative societal impacts of the work performed?
    \item[] Answer: \answerYes{} % Replace by \answerYes{}, \answerNo{}, or \answerNA{}.
    \item[] Justification: %\justificationTODO{}
    We discuss both potential positive societal impacts and negative societal impacts of the work performed in the conclusion section.
    
\item {\bf Safeguards}
    \item[] Question: Does the paper describe safeguards that have been put in place for responsible release of data or models that have a high risk for misuse (e.g., pretrained language models, image generators, or scraped datasets)?
    \item[] Answer: \answerNA{} % Replace by \answerYes{}, \answerNo{}, or \answerNA{}.
    \item[] Justification: %\justificationTODO{}
    This work does not involve such risks.

\item {\bf Licenses for existing assets}
    \item[] Question: Are the creators or original owners of assets (e.g., code, data, models), used in the paper, properly credited and are the license and terms of use explicitly mentioned and properly respected?
    \item[] Answer: \answerYes{} % Replace by \answerYes{}, \answerNo{}, or \answerNA{}.
    \item[] Justification: %\justificationTODO{}
   The original papers or websites that produced the code or dataset are properly cited and we use an open-source dataset for our experiments.

\item {\bf New assets}
    \item[] Question: Are new assets introduced in the paper well documented and is the documentation provided alongside the assets?
    \item[] Answer: \answerYes{} % Replace by \answerYes{}, \answerNo{}, or \answerNA{}.
    \item[] Justification: %\justificationTODO{}
    The code in the paper is well documented and the documentation is provided
alongside the code.

\item {\bf Crowdsourcing and research with human subjects}
    \item[] Question: For crowdsourcing experiments and research with human subjects, does the paper include the full text of instructions given to participants and screenshots, if applicable, as well as details about compensation (if any)? 
    \item[] Answer: \answerNo{} % Replace by \answerYes{}, \answerNo{}, or \answerNA{}.
    \item[] Justification: %\justificationTODO{}
    The paper has no crowdsourcing nor research with human subjects.

\item {\bf Institutional review board (IRB) approvals or equivalent for research with human subjects}
    \item[] Question: Does the paper describe potential risks incurred by study participants, whether such risks were disclosed to the subjects, and whether Institutional Review Board (IRB) approvals (or an equivalent approval/review based on the requirements of your country or institution) were obtained?
    \item[] Answer: \answerNA{} % Replace by \answerYes{}, \answerNo{}, or \answerNA{}.
    \item[] Justification: %\justificationTODO{}
     The paper does not involve crowdsourcing nor research with
human subjects.

\item {\bf Declaration of LLM usage}
    \item[] Question: Does the paper describe the usage of LLMs if it is an important, original, or non-standard component of the core methods in this research? Note that if the LLM is used only for writing, editing, or formatting purposes and does not impact the core methodology, scientific rigorousness, or originality of the research, declaration is not required.
    %this research? 
    \item[] Answer: \answerNA{} % Replace by \answerYes{}, \answerNo{}, or \answerNA{}.
    \item[] Justification: %\justificationTODO{}
    The core method development in this research does not involve LLMs as any important, original, or non-standard components.

\end{enumerate}

%% file: appendix.tex
% \section*{Appendix}

\appendix

\section{Global state cache}
\label{sec:global_state_cache}
In this section, we clarify the design of the global state cache. 
It is composed of a current cache and a previous cache. 
The current one is used to store the predicted sub-images of $T_{12i:12(i+1)}$, and the previous one, storing sub-images of $T_{12(i-1):12i}$ is used to provide neighbour information for the prediction.
\begin{python}
class GlobalStateCache:
    def __init__(self, domains, domain_Chs, time_window,):
        self.domains = domains
        self.time_window = time_window
        self.domain_Chs = domain_Chs

        self.cur_cache = self.init_cache()
        self.prev_cache = self.init_cache()

    def init_cache(self):
        domain_cache = {}
        for domain in self.domains:
            domain_Ch = self.domain_Chs[domain]
            # original image size is 1152x2304.
            # 8x16 subimages with the size of 144x144.
            # 9x9 is the number of tokens in a subimage. 
            domain_cache[domain] = /
            torch.zeros((8, 16, self.time_window, domain_Ch, 9, 9))
        return domain_cache
\end{python}

The main operators of the global state cache are query neighbors and update cache. 
Query neighbors is used to get the adjacent sub-images as conditions for the prediction of central areas. 
Update cache is used to update the previous and current cache with the subsequent 12 hours predictions.
\begin{python}
    def query_neighbours(self, rel_coordinate):
        neighbour_coords = self._get_8_neighbour_coord(rel_coordinate)
        neighbours = {}
        for domain in self.domains:
            neighbours[domain] = []
            for coord in neighbour_coords:
                neighbours[domain].append(
                    self.prev_cache[domain][coord[0], coord[1]]
                    )
        return neighbours

    def update_cache(self, rel_coordinate, pred_subimg):
        for domain in self.domains:
            coords = rel_coordinate[0], rel_coordinate[1]
            self.cur_cache[domain][coords] = pred_subimg[domain]
        if have_pred_whole_img:
            self.prev_cache = self.cur_cache
        return None
\end{python}

To get the neighbour coordinates from a Plane Rectangular Coordinate System, we utilized a \_get\_8\_neighbour\_coord function that incorporates Earth's spherical geometry, 
specifically handling the left-right and top-bottom boundaries of the image.
\begin{python}
    def _get_8_neighbour_coord(self, rel_coordinate, h=8, w=16):
        """
        h, w is the height and width of the image
        r, c is the coordinate of the pixel
        case 1: if the pixel is in the center of the image, return all 8 neighbours
        case 2: if the pixel in w border, treat the image as h border is connected
        case 3: if the pixel in h border, symmetrically get the neighour
        case 4: if the pixel in corner, use the rule of both w border and h border 
        ret:
        8 neighbours ordered as [up_left, up, up_right, left, right, down_left, down, down_right] 
        """
        r, c = rel_coordinate[0], rel_coordinate[1]
        assert r >= 0 and r < h
        assert c >= 0 and c < w
        w_border_flag = (c == 0 or c == w - 1)
        h_top_border_flag = (r == 0)
        h_bottom_border_flag = (r == (h - 1))

        neighbours = []
        if not (h_top_border_flag or h_bottom_border_flag):
            up_left = ((r - 1) % h, (c - 1) % w)
            up = ((r - 1) % h, c)
            up_right = ((r - 1) % h, (c + 1) % w)
            left = (r, (c - 1) % w)
            right = (r, (c + 1) % w)
            down_left = ((r + 1) % h, (c - 1) % w)
            down = ((r + 1) % h, c)
            down_right = ((r + 1) % h, (c + 1) % w)
        elif h_top_border_flag:
            up_left = (r, (c + 1 + w//2) % w)
            up = (r, (c + w // 2)% w)
            up_right = (r, (c - 1 + w//2) % w)
            left = (r, (c - 1) % w)
            right = (r, (c + 1) % w)
            down_left = ((r + 1) % h, (c - 1) % w)
            down = ((r + 1) % h, c)
            down_right = ((r + 1) % h, (c + 1) % w)
        elif h_bottom_border_flag:
            up_left = ((r - 1) % h, (c - 1) % w)
            up = ((r - 1) % h, (c)% w)
            up_right = ((r - 1) % h, (c + 1) % w)
            left = (r, (c - 1) % w)
            right = (r, (c + 1) % w)
            down_left = (r, (c + 1 + w//2) % w)
            down = (r, (c + w//2)% w)
            down_right = (r, (c - 1 + w//2) % w)
        else:
            raise NotImplementedError

        neighbours.append(up_left)
        neighbours.append(up)
        neighbours.append(up_right)
        neighbours.append(left)
        neighbours.append(right)
        neighbours.append(down_left)
        neighbours.append(down)
        neighbours.append(down_right)
        return neighbours
\end{python}

\section{Artificial intelligence data assimilation}
% ask kun % aida + aiwp % pure aida % aida + aidop
% data assimilation influenced 
% observation space -> physical space % 改变模式或者算法 4dvar diffda adas
% observation space -> observation space earthnet
The development of data assimilation has also been revolutionized by artificial intelligence. 
Xiao et al.~\cite{xiao2024towards} were the first to apply the popular traditional numerical data assimilation method, Four-Dimensional Variational, to the AIWP model FengWu.  
Furthermore, researchers have explored the development of artificial intelligence assimilation methods, such as FNP~\cite{chen2024fnp} and DiffDA~\cite{huang2024diffda}, which could be applied to both NWP and AIWP models.  
Although impressive progress has been made, these methods remain limited by reanalysis data and NWP models, which require transforming the observations into physical space.
Unlike previous methods, EarthNet~\cite{vandal2024global} proposes implementing observation space data assimilation with masked reconstruction. 
We are motivated to use an observation AIDA model for formulating a complete observation space.  

\section{Comparisons with spatiotemporal learning methods}

\input{Tables/sup_dop}
As shown in Table~\ref{tab:sub_dop}, we compare DAWP with more spatiotemporal methods including RNN-based (~\cite{shi2015convolutional}, ~\cite{wang2022predrnn}), CNN-based (~\cite{gao2022simvp}, ~\cite{tan2023temporal}), and transformer-based (~\cite{bai2022rainformer}, ~\cite{gao2022earthformer}). 
Our DAWP maintains a significant advantage over these methods, demonstrating the effectiveness of our AIDA module in improving the roll-out and efficiency of AIWP. 

We present implementation details of EarthNet and Transformer-DOP here. There is no open-sourced code for EarthNet~\cite{vandal2024global} or Transformer-DOP~\cite{mcnally2024data}. For EarthNet~\cite{vandal2024global}, it follows the implementation of MultiMAE~\cite{bachmann2022multimae} as detailed in EarthNet's Appendix C and D. Therefore, we reproduce it on our datasets following MultiMAE. As for Transformer-DOP, since the original paper presents only a sketch without details, we implemented it according to our best available understanding. Specifically, EarthNet is reproduced as a 12-layer encoder (hidden dimension 768) paired with an 8-layer decoder (hidden dimension 512), and Transformer-DOP is implemented as a transformer consisting of 18 layers (hidden dimension 1024). We employ sub-images because the full 12-hour global observation sequence would result in 124k-token sequence, which is computationally infeasible.  

\section{VAE comparison}
\label{sec:vaes_comparison}
\input{Tables/vae_table}

% setting: 对比SD-VAE在压缩多channel有缺失观测的satellite数据的能力 
% 介绍表
% multichannel 上优秀, 少channel时compare
% mask 对稀疏观测的影响
We explore the ability of our mask ViT-VAE to compress satellite data with multiple channels and missing values by comparing it with other VAEs.

The results are shown in Table~\ref{tab:vae}, where we compare the reconstruction loss of different VAEs.
First, VAEs with ViT structure are more effective for reconstructing modality data with multiple channels, such as HIRS and AMSU-A, while on modalities with fewer channels, there is only a slight increase in reconstruction loss.
Another observation is that the application of a mask consistently increases VAEs' reconstruction ability on HIRS.
For ViT-VAE, it is beneficial to use the mask for the computation of attention between patch tokens, as it could directly weaken the influence of missing tokens.  

\section{Satellite data preprocessing}
\label{sec:data_preproc}
\textbf{Preprocessing}: The original satellite observation data points are extremely sparse and irregular. To spatially align different observation sources and channels for model training, a remapping procedure is performed beforehand. The pseudocode for the preprocessing algorithm is given in algorithm~\ref{alg:remap}.

\begin{algorithm}
    \label{alg:remap}
	\caption{Remapping Satellite Observation}\label{algorithm}
	\KwIn{target resolution $R$, observation $D_{o}$, corresponding latitudes $C_{lat}$ and longitudes $C_{lon}$}
	\KwOut{remapped observation data points $D_{grid}$ on desired global grid}
	Generate global grid $C_{R}$ of desired resolution $R$ that follows Equirectangular projection\;
	Assign latitudes and longitudes ($C_{lat}$,$C_{lon}$) to the nearest coordinates ($C^{\prime}_{lat}$,$C^{\prime}_{lon}$) on grid $C_{R}$\;
	$D_{grid}\leftarrow$ NaN with the shape of $C_{R}$\;
        $D_{count}\leftarrow$ Zeros with the shape of $C_{R}$\;
	\For{$d_{o}$,$c^{\prime}_{lat}$,$c^{\prime}_{lon}$  in $D_{o}$,$C^{\prime}_{lat}$,$C^{\prime}_{lon}$}
		{
			Locate corresponding data point $d_{grid}$ of $D_{grid}$ according to coordinates ($c^{\prime}_{lat}$,$c^{\prime}_{lon}$)\;
                Locate corresponding point counter $d_{count}$ of $D_{count}$ according to coordinates ($c^{\prime}_{lat}$,$c^{\prime}_{lon}$)\;
			\eIf{$d_{grid}$ equals NaN}
				{$d_{grid} \leftarrow d_{o}$\;}
				{$d_{grid} \leftarrow (d_{grid} + d_{o})$\;}
		}
        Average each $d_{grid}$ where $d_{count} \ge 1$
\end{algorithm}

\begin{figure}[t]
    \centering
    \includegraphics[width=1.0\linewidth]{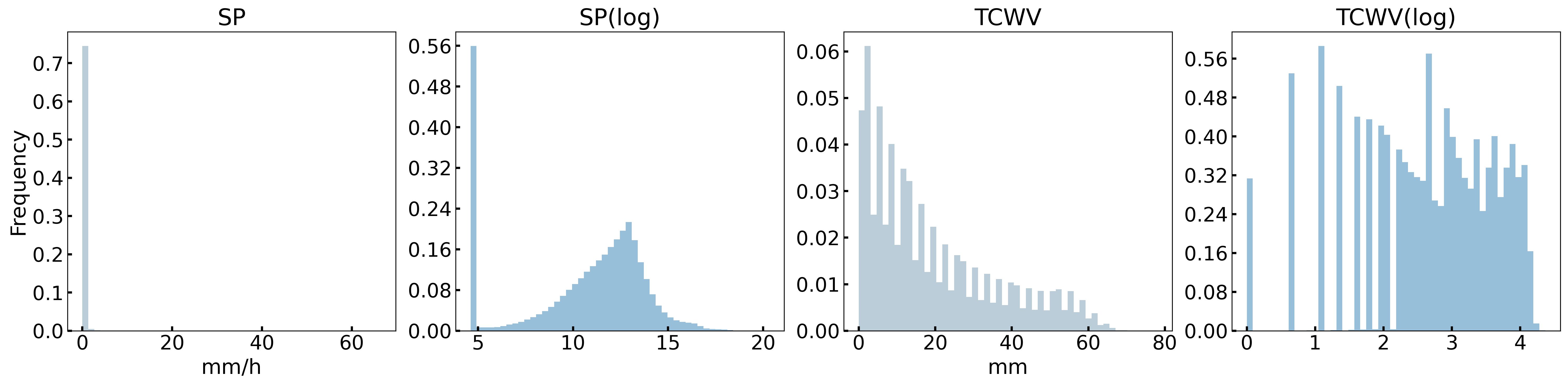}
    \vspace{-0.7cm}
    \caption{Distribution of ATMS precipitation productions. SP(log) indicates applying a log-transform on SP. It is the same for TCWV(log).}
    \label{fig:prec_dist}
\end{figure}

\textbf{Normalization}: We normalize each modality for efficient convergence. The direct observation modalities are normalized by:
\begin{equation}
    x_{norm}^{M} = \frac{x^M -Mean({x^M})}{Std(x^M)},
\end{equation}
where $M$ represents the modality $M$. 
For ATMS-precipitation, we first implement the log-transformation as: 
\begin{equation}
    x_{prec} = \log{(x/a +b)},
\end{equation}
to alleviate long-tail distribution, and then normalize these variables like other modalities. Specifically, we select $a=1e-7$, $b=1e2$ for SP channel, and choose $a=1$, $b=1$ for TCWV channel. The distribution shift is shown in Figure~\ref{fig:prec_dist}. The original distributions also motivate us to choose a threshold list of [0.5 mm/h, 1 mm/h, 2 mm/h] for SP.

\section{Dataset introduction}
\label{sec:dataset_introduction}
The four sensors below are organized into a satellite observation dataset for direct observation forecasting. 
This composite dataset has a training split with data from January of 2012 to June of 2022 for training and a testing split composed of data from May of 2023 to July of 2023. We present detailed introductions to these sensors. 

\textbf{AMSU-A}: The Advanced Microwave Sounding Unit-A (AMSU-A) is a 15-channel microwave radiometer used for measuring global atmospheric temperature profiles and will provide information on atmospheric water in all of its phases (with the exception of small ice particles, which are transparent at microwave frequencies). AMSU-A will provide information even in cloudy conditions. AMSU-A measures Earth radiance at frequencies (in GHz) as listed under the instrument channel information. Level 1B data was collected from EUMETSAT at \url{https://archive.eumetsat.int/usc/UserServicesClient.html}.

\textbf{ATMS}: The Advanced Technology Microwave Sounder (ATMS) and the Cross-track Infrared Sounder (CrIS) work together to provide global high-resolution profiles of temperature and moisture. These advanced atmospheric sensors create cross-sections of storms and other weather conditions, helping with both short-term nowcasting and long-term forecasting. Level 1C data was collected from GES DISC at \url{https://disc.gsfc.nasa.gov/datasets?page=1}.

\textbf{HIRS}: The High Resolution Infrared Sounder (HIRS) operates at 20 channels (19 channels in the infrared and one in the visible). Its main purpose is to provide input for the vertical temperature and humidity profile retrievals. In addition, the HIRS pixel resolution serves as the standard grid resolution for all ATOVS level 2 products. Level 1B data was collected from EUMETSAT at \url{https://archive.eumetsat.int/usc/UserServicesClient.html}.

\textbf{MHS}: The Microwave Humidity Sounder (MHS) is a 5 channel instrument used to provide input to the retrieval of surface temperatures, emissivities, and atmospheric humidity. In combination with AMSU-A information it can also be used to process precipitation rates and related cloud properties, as well as to detect sea ice and snow coverage. Level 1B data was collected from EUMETSAT at \url{https://archive.eumetsat.int/usc/UserServicesClient.html}.

\textbf{ATMS-Precipitation}: The ATMS-Precipitation is one of the products of the Global Precipitation Measurement (GPM) mission. It is based on the L1C-level calibrated brightness temperature data of the ATMS sensor and extracts information such as precipitation rate and precipitation type through a physical inversion algorithm. Level 2A data was collected from GES DISC at \url{https://disc.gsfc.nasa.gov/datasets?page=1}.

\section{Training details}
Our DAWP framework is trained in 3 stages on 4 A100 80G GPUs, including training mask ViT-VAEs for encoding and mapping, an MMAE for data assimilation in observation space, and a spatiotemporal transformer for direct observation prediction.
Specifically, these modules are all trained within 144$\times$144 sub-images. 
The encoder and decoder of the mask ViT-VAE use the same transformer structure with a patch size of 16 and a hidden dimension of 768. 
It is trained with a reconstruction MAE loss and a KL loss weight of 0.000001 for robust representation.
We freeze the pretrained mask ViT-VAE as the encoders for each modality in our MMAE. 
Each modality observation with a 12h time window in a 144$\times$144 sub-image is tokenized into 972 spatiotemporal tokens. 
Thus, our MMAE totally received 3888 tokens. 
We randomly select 128 observed tokens of them (~3.3\%) to reconstruct the remaining observed tokens via MAE training. 
Given the 144$\times$ 144 sub-images assimilated by MMAE, our spatiotemporal transformer is trained. 
It is structured with 12 TS spatiotemporal decoupling blocks, whose hidden dimension is 768.
The hyperparameters for optimizing these modules are similar. 
All of them use the AdamW optimizer with $\beta_0=0.9$, $\beta_1=0.999$, and a learning rate of 0.0001. 
The learning rate is scheduled by a cosine scheduler, warming up 10k steps, step by step.
\label{sec:training_details}

\input{Tables/training_vae}

\input{Tables/training_aida}

\input{Tables/training_aiwp}

\input{Tables/vae_structure.tex}
\input{Tables/mmae_structure.tex}
\input{Tables/aiwp_structure.tex}

\input{Tables/computation_cost}
In the table ~\ref{tab:comp_cost}, we present the computation cost.

\section{Metrics defination}
\label{sec:metrics_defination}

\subsection{CSI and FAR}
For the evaluation of global precipitation variables, the metrics include the Critical Success Index (CSI) and False Alarm Ratio (FAR). They are core binary classification evaluation metrics that quantify the detection accuracy and reliability of precipitation events.
In the field of meteorology, these metrics assess the consistency and accuracy between precipitation predictions and observed results, quantitatively evaluating the performance of models. 
To measure the accuracy of prediction for precipitation with different intensities. 
Before calculating these metrics, we transform the predicted pixel values and ground truth into binary values (0 or 1) using a given threshold $\tau$.
The value is set to 0 if it is less than 
$\tau$; otherwise, it is set to 1.  
These binary values enable us to determine the true positive (TP), false positive (FP), false negative (FN), and true negative (TN) counts. CSI, HSS, and FSS are calculated by these counts as follows:

1) Critical Success Index.
CSI is a metric that evaluates the proportion of correctly predicted events of hits among conditions, including hits (TP), false alarms (FN), and misses (FP). The formulation of CSI is:
\begin{equation}
    CSI = \frac{TP}{TP+FN+FP}
\end{equation}
The value of CSI ranges from 0 to 1. Higher values indicate better prediction accuracy.

2) False Alarm Ratio.
The FAR metric quantifies the proportion of predicted positive events that were actually negative in meteorological verification, emphasizing the reliability of alarm triggers. It is defined as:
\begin{equation}
    FAR = \frac{FP}{FP+TP}
\end{equation}
where FP denotes false positive predictions (e.g., forecasted rainfall with no ground observation) and TP represents true positives (correctly predicted rainfall events). 
FAR ranges from 0 (perfect reliability) to 1 (all alarms are false), with lower values indicating better prediction specificity.

\subsection{MAE}
To evaluate the accuracy of direct observation predictions, we use a pointwise Mean Absolute Error (MAE) as the metric to calculate errors on the ground truth with variable missing values. It is worth noting that the MAE is calculated with the raw observation point by point to ignore the influence of missing values.
It is defined as:
\begin{equation}
    MAE = \frac{1}{N} \sum_{i=1}^{N} |y_i - \hat{y}_i|
\end{equation}
N is the total number of points with observation, $y_i$ is the ground truth at the $i^{th}$ location, and $\hat{y}_i$ is the prediction.

\section{More results of direct observation predictions}
\label{sec:DOP_results}

\begin{figure}[t]
    \centering
    \includegraphics[width=1.0\linewidth]{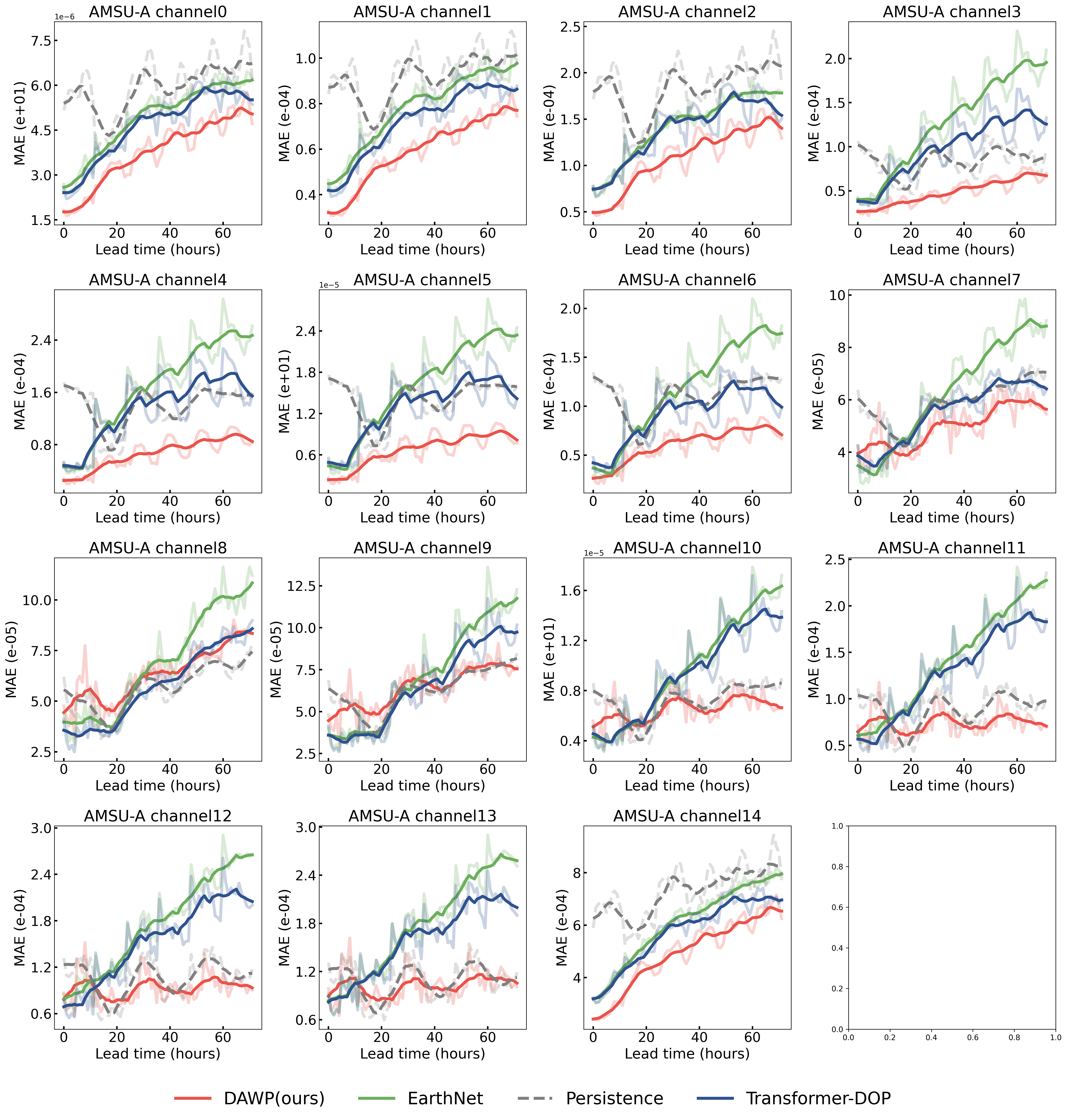}
    \vspace{-0.7cm}
    \caption{Curves of MAE for the prediction of different channels in sensor AMSU-A. The max leadtime is 72h with a 1h temporal resolution.}
\end{figure}

\begin{figure}[t]
    \centering
    \includegraphics[width=1.0\linewidth]{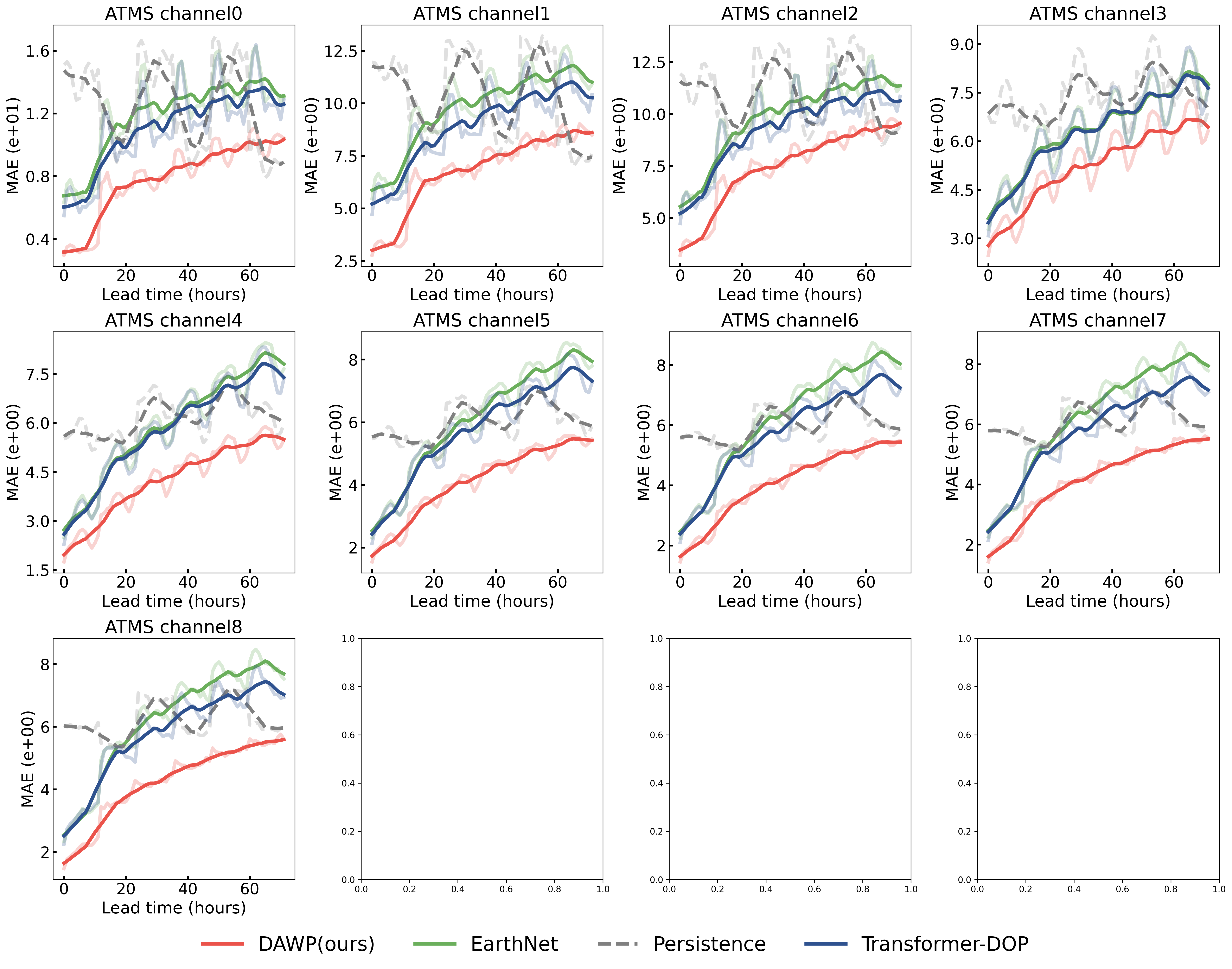}
    \vspace{-0.7cm}
    \caption{Curves of MAE for the prediction of different channels in sensor ATMS. The max leadtime is 72h with a 1h temporal resolution.}
\end{figure}

\begin{figure}[t]
    \centering
    \includegraphics[width=1.0\linewidth]{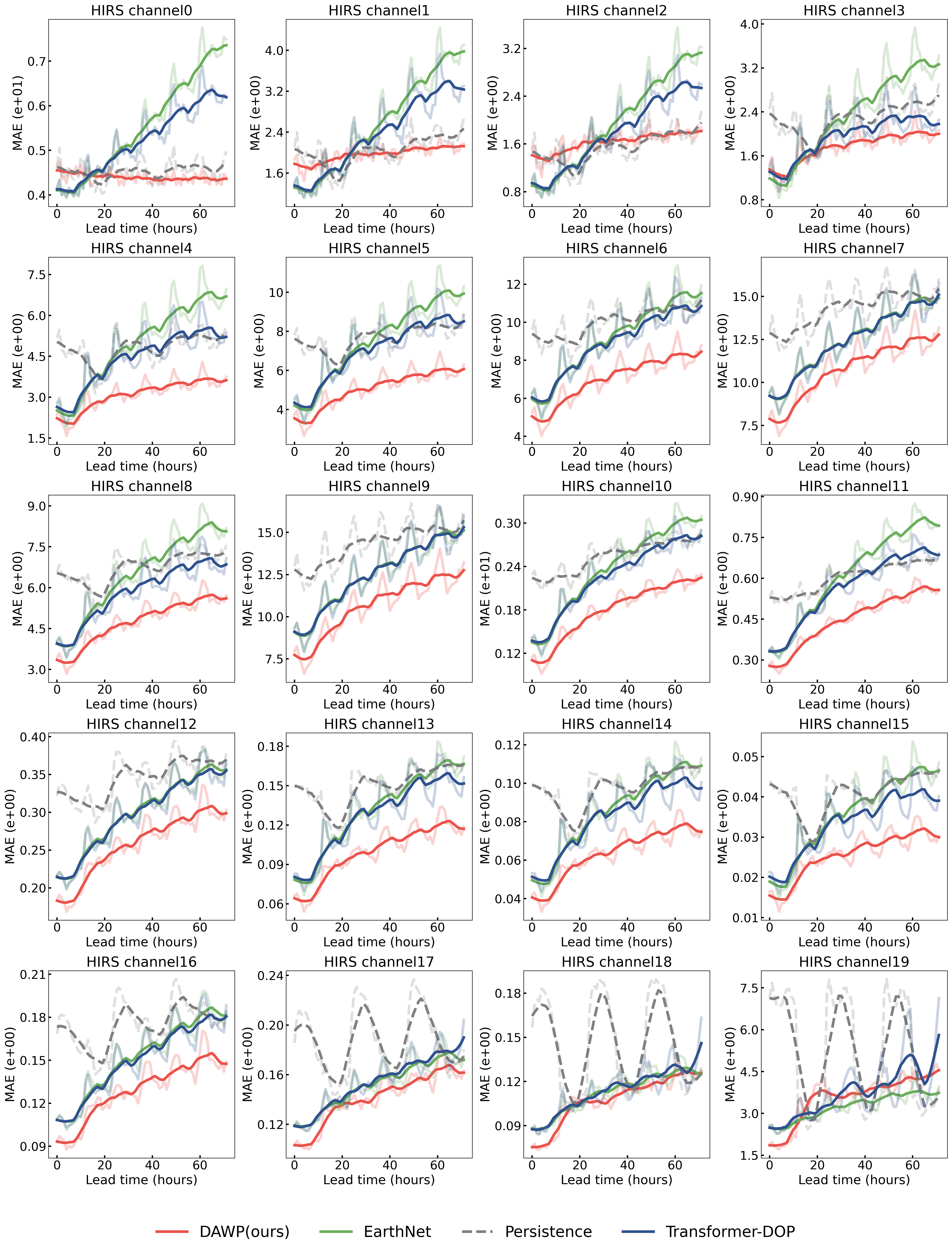}
    \vspace{-0.7cm}
    \caption{Curves of MAE for the prediction of different channels in sensor HIRS. The max leadtime is 72h with a 1h temporal resolution.}
\end{figure}

\begin{figure}[t]
    \centering
    \includegraphics[width=1.0\linewidth]{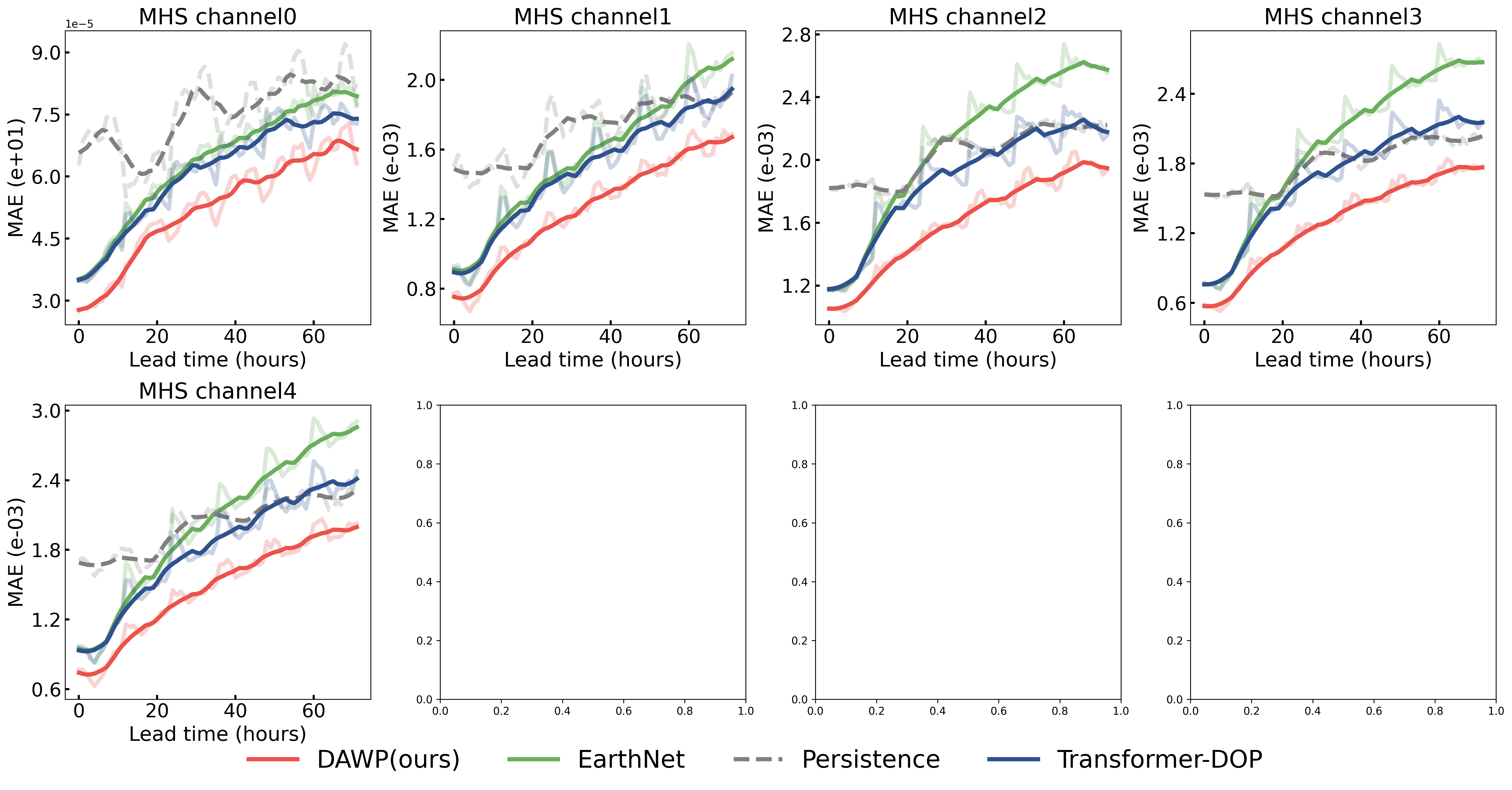}
    \vspace{-0.7cm}
    \caption{Curves of MAE for the prediction of different channels in sensor MHS. The max leadtime is 72h with a 1h temporal resolution.}
\end{figure}

\begin{figure}[t]
    \centering
    \includegraphics[width=1.0\linewidth]{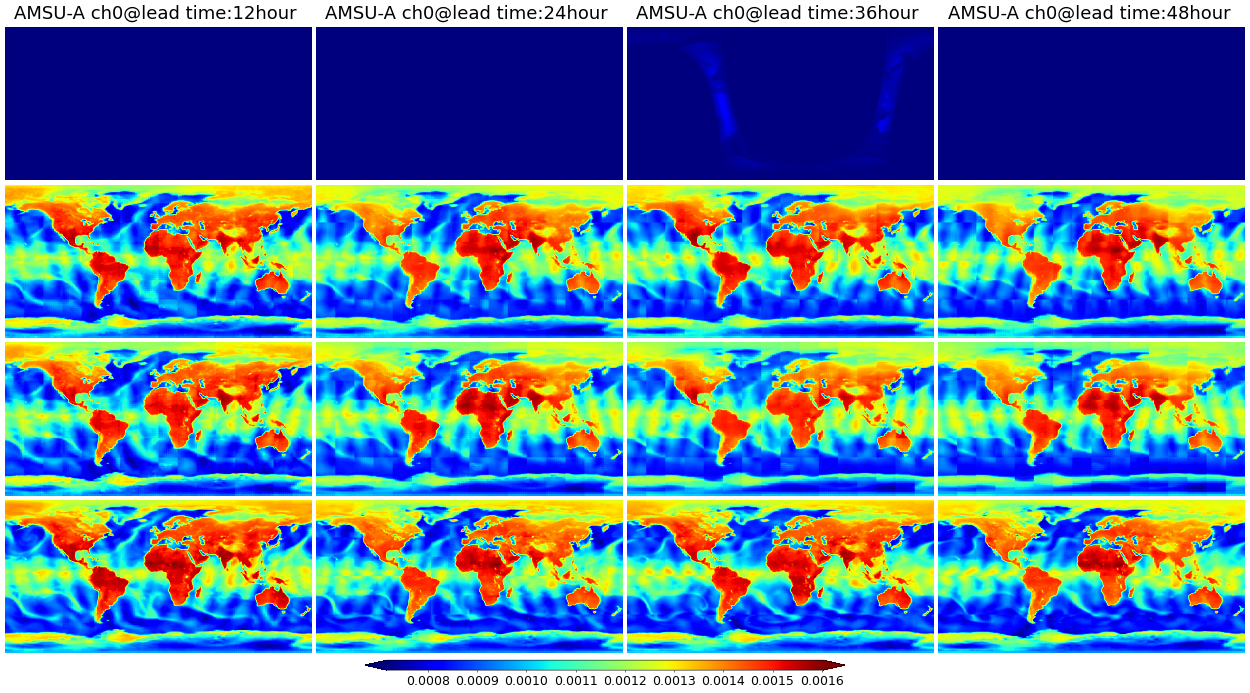}
    \vspace{-0.7cm}
    \caption{A visualization of rollout predictions for channel 0 of AMSU-A. From top to bottom are the results of ground truth, Transformer-DOP, EarthNet, and DAWP (ours).}
\end{figure}

\begin{figure}[t]
    \centering
    \includegraphics[width=1.0\linewidth]{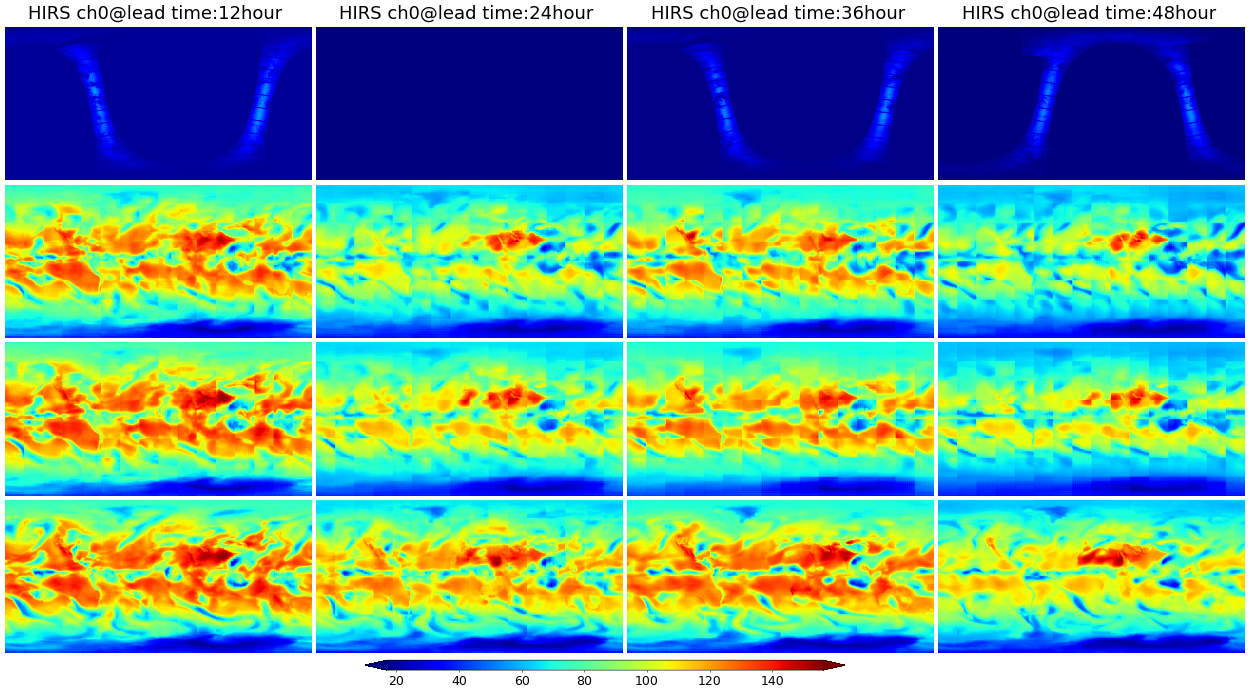}
    \vspace{-0.7cm}
    \caption{A visualization of rollout predictions for channel 9 of HIRS. From top to bottom are the results of ground truth, Transformer-DOP, EarthNet, and DAWP (ours).}
\end{figure}

\begin{figure}[t]
    \centering
    \includegraphics[width=1.0\linewidth]{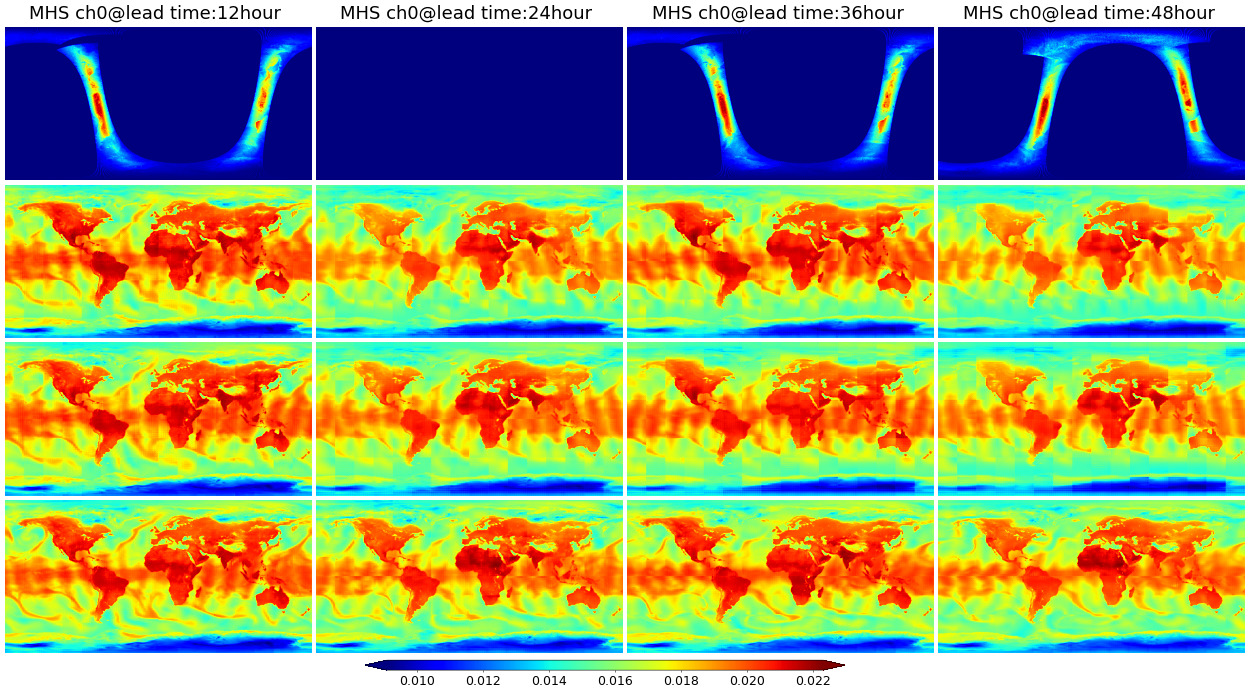}
    \vspace{-0.7cm}
    \caption{A visualization of rollout predictions for channel 0 of MHS. From top to bottom are the results of ground truth, Transformer-DOP, EarthNet, and DAWP (ours).}
\end{figure}

\begin{figure}[t]
    \centering
    \includegraphics[width=1.0\linewidth]{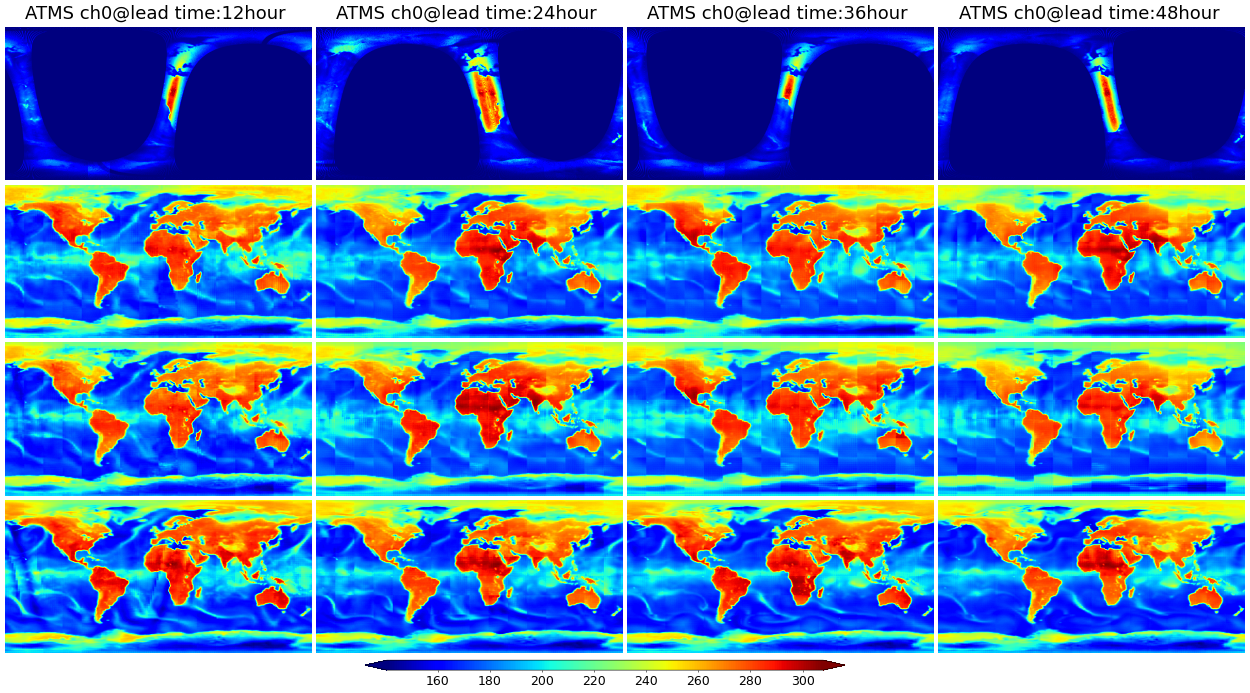}
    \vspace{-0.7cm}
    \caption{A visualization of rollout predictions for channel 1 of ATMS. From top to bottom are the results of ground truth, Transformer-DOP, EarthNet, and DAWP (ours).}
\end{figure}

\begin{figure}[t]
    \centering
    \includegraphics[width=1.0\linewidth]{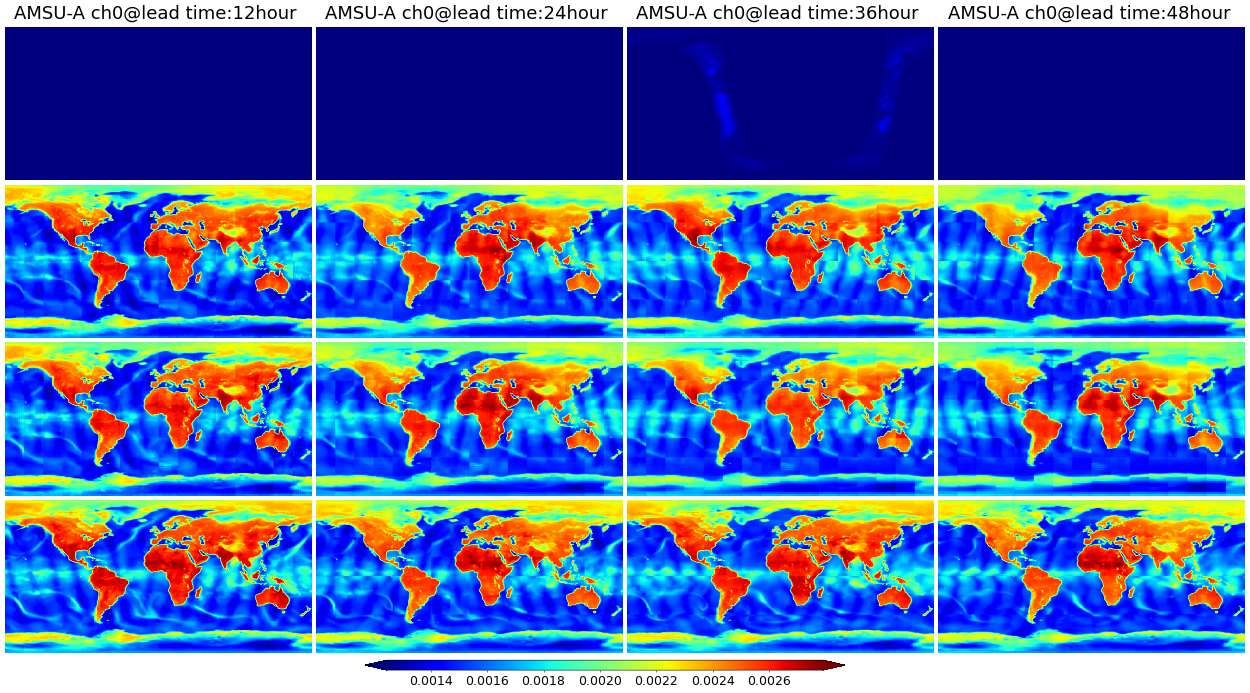}
    \vspace{-0.7cm}
    \caption{A visualization of rollout predictions for channel 1 of AMSU-A. From top to bottom are the results of ground truth, Transformer-DOP, EarthNet, and DAWP (ours).}
\end{figure}

\begin{figure}[t]
    \centering
    \includegraphics[width=1.0\linewidth]{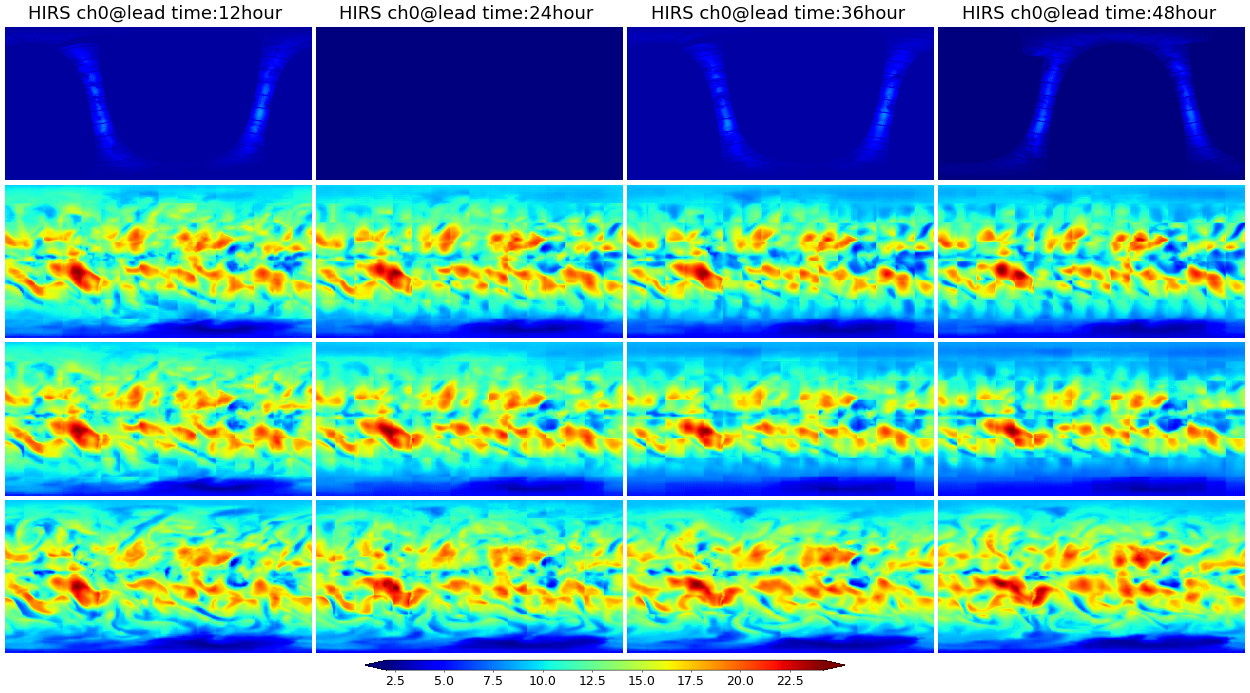}
    \vspace{-0.7cm}
    \caption{A visualization of rollout predictions for channel 10 of HIRS. From top to bottom are the results of ground truth, Transformer-DOP, EarthNet, and DAWP (ours).}
\end{figure}

\begin{figure}[t]
    \centering
    \includegraphics[width=1.0\linewidth]{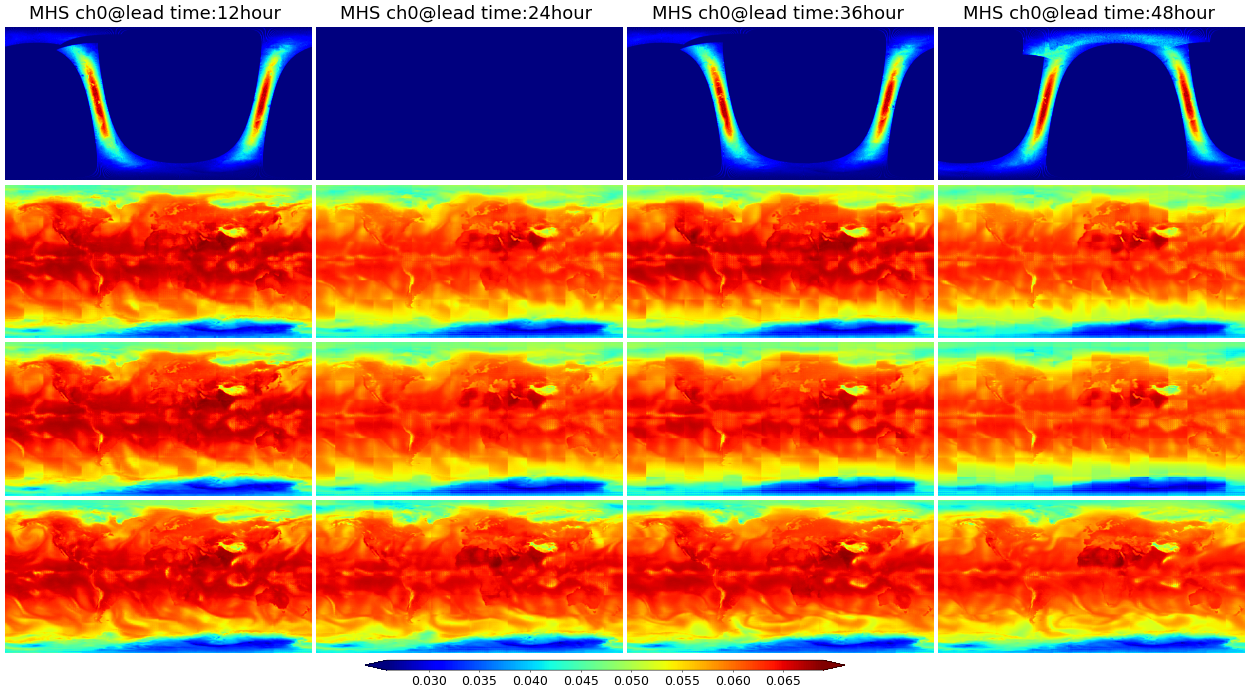}
    \vspace{-0.7cm}
    \caption{A visualization of rollout predictions for channel 1 of MHS. From top to bottom are the results of ground truth, Transformer-DOP, EarthNet, and DAWP (ours).}
\end{figure}

%% file: Tables/sup_dop.tex
% \usepackage{multirow}
\begin{table}[h]
\centering
\caption{More results on spatiotemporal methods. MAE error of forecasting during 3 lead time periods (0-12h, 12-24h, and 24-36h) for different channels of the satellite data. 
We use the unit of 1e-5 for AMSU-A, 1e-4 for MHS, and 1e-0 for both ATMS and HIRS.}
\resizebox{\linewidth}{!}{
\begin{tabular}{l|c|cc|cc|cc|cc}
\toprule
\multicolumn{1}{c|}{}                          &                             & \multicolumn{2}{c|}{AMSU-A}                                      & \multicolumn{2}{c|}{ATMS}                                      & \multicolumn{2}{c|}{HIRS}                                      & \multicolumn{2}{c  }{MHS}                                        \\
\multicolumn{1}{c|}{\multirow{-2}{*}{Methods}} & \multirow{-2}{*}{Lead time} & ch0                            & ch1                            & ch0                           & ch1                           & ch9                           & ch10                          & ch0                           & ch1                            \\ \midrule
Persistence~\cite{gao2022earthformer}                                   &                             & 5.86                           & 9.15                           & 14.37                         & 11.69                         & 12.43                         & 2.21                          & 7.01                          & 14.74                          \\ 
\cellcolor[HTML]{EFEFEF}ConvLSTM~\cite{shi2015convolutional}              &                             & \cellcolor[HTML]{EFEFEF}127.82 & \cellcolor[HTML]{EFEFEF}208.90 & \cellcolor[HTML]{EFEFEF}73.21 & \cellcolor[HTML]{EFEFEF}78.44 & \cellcolor[HTML]{EFEFEF}77.40 & \cellcolor[HTML]{EFEFEF}11.05 & \cellcolor[HTML]{EFEFEF}62.27 & \cellcolor[HTML]{EFEFEF}158.57 \\
\cellcolor[HTML]{EFEFEF}PredRNN~\cite{wang2022predrnn}               &                             & \cellcolor[HTML]{EFEFEF}2.95   & \cellcolor[HTML]{EFEFEF}4.96   & \cellcolor[HTML]{EFEFEF}6.99  & \cellcolor[HTML]{EFEFEF}6.21  & \cellcolor[HTML]{EFEFEF}9.26  & \cellcolor[HTML]{EFEFEF}1.42  & \cellcolor[HTML]{EFEFEF}4.11  & \cellcolor[HTML]{EFEFEF}10.20  \\
\cellcolor[HTML]{EFEFEF}RainFormer~\cite{bai2022rainformer}            &                             & \cellcolor[HTML]{EFEFEF}3.95   & \cellcolor[HTML]{EFEFEF}6.52   & \cellcolor[HTML]{EFEFEF}9.08  & \cellcolor[HTML]{EFEFEF}8.05  & \cellcolor[HTML]{EFEFEF}10.41 & \cellcolor[HTML]{EFEFEF}1.63  & \cellcolor[HTML]{EFEFEF}4.98  & \cellcolor[HTML]{EFEFEF}11.69  \\
\cellcolor[HTML]{EFEFEF}EarthFormer~\cite{gao2022earthformer}           &                             & \cellcolor[HTML]{EFEFEF}18.97  & \cellcolor[HTML]{EFEFEF}33.94  & \cellcolor[HTML]{EFEFEF}37.33 & \cellcolor[HTML]{EFEFEF}38.75 & \cellcolor[HTML]{EFEFEF}22.49 & \cellcolor[HTML]{EFEFEF}3.62  & \cellcolor[HTML]{EFEFEF}18.00 & \cellcolor[HTML]{EFEFEF}55.04  \\
\cellcolor[HTML]{EFEFEF}SimVP~\cite{gao2022simvp}                 &                             & \cellcolor[HTML]{EFEFEF}3.61   & \cellcolor[HTML]{EFEFEF}6.02   & \cellcolor[HTML]{EFEFEF}7.29  & \cellcolor[HTML]{EFEFEF}6.61  & \cellcolor[HTML]{EFEFEF}9.84  & \cellcolor[HTML]{EFEFEF}1.53  & \cellcolor[HTML]{EFEFEF}4.68  & \cellcolor[HTML]{EFEFEF}11.35  \\
\cellcolor[HTML]{EFEFEF}TAU~\cite{tan2023temporal}                   &                             & \cellcolor[HTML]{EFEFEF}3.84   & \cellcolor[HTML]{EFEFEF}6.31   & \cellcolor[HTML]{EFEFEF}7.70  & \cellcolor[HTML]{EFEFEF}6.86  & \cellcolor[HTML]{EFEFEF}9.94  & \cellcolor[HTML]{EFEFEF}1.58  & \cellcolor[HTML]{EFEFEF}4.87  & \cellcolor[HTML]{EFEFEF}11.42  \\
\cellcolor[HTML]{E5E5E5}EarthNet~\cite{vandal2024global}              &                             & \cellcolor[HTML]{E5E5E5}2.93   & \cellcolor[HTML]{E5E5E5}4.89   & \cellcolor[HTML]{E5E5E5}6.96  & \cellcolor[HTML]{E5E5E5}6.14  & \cellcolor[HTML]{E5E5E5}9.15  & \cellcolor[HTML]{E5E5E5}1.39  & \cellcolor[HTML]{E5E5E5}3.96  & \cellcolor[HTML]{E5E5E5}9.65   \\
\cellcolor[HTML]{E5E5E5}Transformer-DOP~\cite{mcnally2024data}       &                             & \cellcolor[HTML]{E5E5E5}2.67   & \cellcolor[HTML]{E5E5E5}4.48   & \cellcolor[HTML]{E5E5E5}6.40  & \cellcolor[HTML]{E5E5E5}5.61  & \cellcolor[HTML]{E5E5E5}9.22  & \cellcolor[HTML]{E5E5E5}1.42  & \cellcolor[HTML]{E5E5E5}3.91  & \cellcolor[HTML]{E5E5E5}9.48   \\
\cellcolor[HTML]{CBCEFB}Ours                  & \multirow{-10}{*}{0-12h}    & \cellcolor[HTML]{CBCEFB}1.92   & \cellcolor[HTML]{CBCEFB}3.39   & \cellcolor[HTML]{CBCEFB}3.36  & \cellcolor[HTML]{CBCEFB}3.27  & \cellcolor[HTML]{CBCEFB}7.70  & \cellcolor[HTML]{CBCEFB}1.12  & \cellcolor[HTML]{CBCEFB}3.07  & \cellcolor[HTML]{CBCEFB}7.91   \\ \midrule
Persistence~\cite{gao2022earthformer}                                    &                             & 4.35                           & 6.94                           & 10.40                         & 8.86                          & 13.58                         & 2.30                          & 6.07                          & 15.09                          \\
\cellcolor[HTML]{EFEFEF}ConvLSTM~\cite{shi2015convolutional}              &                             & \cellcolor[HTML]{EFEFEF}128.13 & \cellcolor[HTML]{EFEFEF}211.14 & \cellcolor[HTML]{EFEFEF}81.22 & \cellcolor[HTML]{EFEFEF}82.60 & \cellcolor[HTML]{EFEFEF}71.64 & \cellcolor[HTML]{EFEFEF}9.98  & \cellcolor[HTML]{EFEFEF}59.26 & \cellcolor[HTML]{EFEFEF}145.44 \\
\cellcolor[HTML]{EFEFEF}PredRNN~\cite{wang2022predrnn}               &                             & \cellcolor[HTML]{EFEFEF}3.85   & \cellcolor[HTML]{EFEFEF}6.26   & \cellcolor[HTML]{EFEFEF}11.14 & \cellcolor[HTML]{EFEFEF}8.82  & \cellcolor[HTML]{EFEFEF}10.92 & \cellcolor[HTML]{EFEFEF}1.84  & \cellcolor[HTML]{EFEFEF}5.48  & \cellcolor[HTML]{EFEFEF}13.16  \\
\cellcolor[HTML]{EFEFEF}RainFormer~\cite{bai2022rainformer}           &                             & \cellcolor[HTML]{EFEFEF}11.46  & \cellcolor[HTML]{EFEFEF}15.72  & \cellcolor[HTML]{EFEFEF}19.43 & \cellcolor[HTML]{EFEFEF}17.76 & \cellcolor[HTML]{EFEFEF}18.75 & \cellcolor[HTML]{EFEFEF}3.28  & \cellcolor[HTML]{EFEFEF}11.74 & \cellcolor[HTML]{EFEFEF}32.10  \\
\cellcolor[HTML]{EFEFEF}EarthFormer~\cite{gao2022earthformer}            &                             & \cellcolor[HTML]{EFEFEF}18.98  & \cellcolor[HTML]{EFEFEF}33.96  & \cellcolor[HTML]{EFEFEF}37.33 & \cellcolor[HTML]{EFEFEF}38.76 & \cellcolor[HTML]{EFEFEF}22.50 & \cellcolor[HTML]{EFEFEF}3.62  & \cellcolor[HTML]{EFEFEF}18.01 & \cellcolor[HTML]{EFEFEF}55.03  \\
\cellcolor[HTML]{EFEFEF}SimVP~\cite{gao2022simvp}                 &                             & \cellcolor[HTML]{EFEFEF}4.83   & \cellcolor[HTML]{EFEFEF}7.74   & \cellcolor[HTML]{EFEFEF}11.48 & \cellcolor[HTML]{EFEFEF}9.10  & \cellcolor[HTML]{EFEFEF}11.49 & \cellcolor[HTML]{EFEFEF}2.02  & \cellcolor[HTML]{EFEFEF}6.43  & \cellcolor[HTML]{EFEFEF}14.20  \\
\cellcolor[HTML]{EFEFEF}TAU~\cite{tan2023temporal}                   &                             & \cellcolor[HTML]{EFEFEF}4.46   & \cellcolor[HTML]{EFEFEF}7.21   & \cellcolor[HTML]{EFEFEF}11.46 & \cellcolor[HTML]{EFEFEF}9.06  & \cellcolor[HTML]{EFEFEF}11.65 & \cellcolor[HTML]{EFEFEF}2.04  & \cellcolor[HTML]{EFEFEF}6.24  & \cellcolor[HTML]{EFEFEF}13.94  \\
\cellcolor[HTML]{E5E5E5}EarthNet~\cite{vandal2024global}             &                             & \cellcolor[HTML]{E5E5E5}4.12   & \cellcolor[HTML]{E5E5E5}6.65   & \cellcolor[HTML]{E5E5E5}11.25 & \cellcolor[HTML]{E5E5E5}9.00  & \cellcolor[HTML]{E5E5E5}11.14 & \cellcolor[HTML]{E5E5E5}1.98  & \cellcolor[HTML]{E5E5E5}5.46  & \cellcolor[HTML]{E5E5E5}13.11  \\
\cellcolor[HTML]{E5E5E5}Transformer-DOP~\cite{mcnally2024data}       &                             & \cellcolor[HTML]{E5E5E5}3.84   & \cellcolor[HTML]{E5E5E5}6.14   & \cellcolor[HTML]{E5E5E5}10.04 & \cellcolor[HTML]{E5E5E5}8.04  & \cellcolor[HTML]{E5E5E5}11.08 & \cellcolor[HTML]{E5E5E5}1.95  & \cellcolor[HTML]{E5E5E5}5.19  & \cellcolor[HTML]{E5E5E5}12.65  \\
\cellcolor[HTML]{CBCEFB}Ours       ~\cite{gao2022earthformer}            & \multirow{-10}{*}{12-24h}   & \cellcolor[HTML]{CBCEFB}3.11   & \cellcolor[HTML]{CBCEFB}5.12   & \cellcolor[HTML]{CBCEFB}7.35  & \cellcolor[HTML]{CBCEFB}6.35  & \cellcolor[HTML]{CBCEFB}9.57  & \cellcolor[HTML]{CBCEFB}1.54  & \cellcolor[HTML]{CBCEFB}4.51  & \cellcolor[HTML]{CBCEFB}10.54  \\ \midrule
Persistence~\cite{gao2022earthformer}                                 &                             & 6.39                           & 9.84                           & 15.37                         & 12.52                         & 14.61                         & 2.61                          & 8.00                          & 17.86                          \\
\cellcolor[HTML]{EFEFEF}ConvLSTM~\cite{shi2015convolutional}         &                             & \cellcolor[HTML]{EFEFEF}128.42 & \cellcolor[HTML]{EFEFEF}211.83 & \cellcolor[HTML]{EFEFEF}82.03 & \cellcolor[HTML]{EFEFEF}85.17 & \cellcolor[HTML]{EFEFEF}72.62 & \cellcolor[HTML]{EFEFEF}10.15 & \cellcolor[HTML]{EFEFEF}60.44 & \cellcolor[HTML]{EFEFEF}148.62 \\
\cellcolor[HTML]{EFEFEF}PredRNN~\cite{wang2022predrnn}               &                             & \cellcolor[HTML]{EFEFEF}4.58   & \cellcolor[HTML]{EFEFEF}7.23   & \cellcolor[HTML]{EFEFEF}12.18 & \cellcolor[HTML]{EFEFEF}9.69  & \cellcolor[HTML]{EFEFEF}12.03 & \cellcolor[HTML]{EFEFEF}2.08  & \cellcolor[HTML]{EFEFEF}6.24  & \cellcolor[HTML]{EFEFEF}15.04  \\
\cellcolor[HTML]{EFEFEF}RainFormer~\cite{bai2022rainformer}            &                             & \cellcolor[HTML]{EFEFEF}24.89  & \cellcolor[HTML]{EFEFEF}32.34  & \cellcolor[HTML]{EFEFEF}35.49 & \cellcolor[HTML]{EFEFEF}36.54 & \cellcolor[HTML]{EFEFEF}30.78 & \cellcolor[HTML]{EFEFEF}4.92  & \cellcolor[HTML]{EFEFEF}21.51 & \cellcolor[HTML]{EFEFEF}69.68  \\
\cellcolor[HTML]{EFEFEF}EarthFormer~\cite{gao2022earthformer}          &                             & \cellcolor[HTML]{EFEFEF}18.98  & \cellcolor[HTML]{EFEFEF}33.96  & \cellcolor[HTML]{EFEFEF}37.34 & \cellcolor[HTML]{EFEFEF}38.77 & \cellcolor[HTML]{EFEFEF}22.51 & \cellcolor[HTML]{EFEFEF}3.63  & \cellcolor[HTML]{EFEFEF}18.02 & \cellcolor[HTML]{EFEFEF}55.04  \\
\cellcolor[HTML]{EFEFEF}SimVP~\cite{gao2022simvp}                 &                             & \cellcolor[HTML]{EFEFEF}5.72   & \cellcolor[HTML]{EFEFEF}8.91   & \cellcolor[HTML]{EFEFEF}12.93 & \cellcolor[HTML]{EFEFEF}10.33 & \cellcolor[HTML]{EFEFEF}12.71 & \cellcolor[HTML]{EFEFEF}2.32  & \cellcolor[HTML]{EFEFEF}73.95 & \cellcolor[HTML]{EFEFEF}15.99  \\
\cellcolor[HTML]{EFEFEF}TAU~\cite{tan2023temporal}                   &                             & \cellcolor[HTML]{EFEFEF}5.61   & \cellcolor[HTML]{EFEFEF}8.76   & \cellcolor[HTML]{EFEFEF}13.23 & \cellcolor[HTML]{EFEFEF}10.53 & \cellcolor[HTML]{EFEFEF}12.98 & \cellcolor[HTML]{EFEFEF}2.34  & \cellcolor[HTML]{EFEFEF}7.15  & \cellcolor[HTML]{EFEFEF}15.85  \\
\cellcolor[HTML]{E5E5E5}EarthNet~\cite{vandal2024global}               &                             & \cellcolor[HTML]{E5E5E5}5.17   & \cellcolor[HTML]{E5E5E5}8.14   & \cellcolor[HTML]{E5E5E5}12.52 & \cellcolor[HTML]{E5E5E5}10.08 & \cellcolor[HTML]{E5E5E5}12.37 & \cellcolor[HTML]{E5E5E5}2.36  & \cellcolor[HTML]{E5E5E5}6.41  & \cellcolor[HTML]{E5E5E5}15.08  \\
\cellcolor[HTML]{E5E5E5}Transformer-DOP~\cite{mcnally2024data}       &                             & \cellcolor[HTML]{E5E5E5}4.91   & \cellcolor[HTML]{E5E5E5}7.54   & \cellcolor[HTML]{E5E5E5}11.35 & \cellcolor[HTML]{E5E5E5}9.07  & \cellcolor[HTML]{E5E5E5}12.39 & \cellcolor[HTML]{E5E5E5}2.27  & \cellcolor[HTML]{E5E5E5}6.22  & \cellcolor[HTML]{E5E5E5}14.70  \\
\cellcolor[HTML]{CBCEFB}Ours                  & \multirow{-10}{*}{24-36h}   & \cellcolor[HTML]{CBCEFB}3.66   & \cellcolor[HTML]{CBCEFB}5.80   & \cellcolor[HTML]{CBCEFB}7.84  & \cellcolor[HTML]{CBCEFB}6.81  & \cellcolor[HTML]{CBCEFB}10.71 & \cellcolor[HTML]{CBCEFB}1.79  & \cellcolor[HTML]{CBCEFB}5.15  & \cellcolor[HTML]{CBCEFB}12.22  \\ \bottomrule
\end{tabular}
}
\label{tab:sub_dop}
\end{table}

%% file: Tables/vae_table.tex
% \usepackage{multirow}
\begin{table}[t]
\caption{Reconstruction error of various VAEs on different modalities. 
The column of \textbf{Improvement} represents relative average improvement over SD-VAE.}
\centering
\resizebox{\linewidth}{!}{
\begin{tabular}{l|cccc|c}
\hline
                       & \multicolumn{4}{c|}{Modality reconstruction error}                                 &                                 \\ \cline{2-5}
\multirow{-2}{*}{VAEs} & AMSU-A (1e-3) & ATMS (1e-2)  & HIRS (1e-3)                          & MHS (1e-2)   & \multirow{-2}{*}{Improvement}   \\ \hline
\rowcolor[HTML]{EFEFEF} 
SD-VAE~\cite{rombach2022high}                 & 1.07          & 1.26         & 5.84                                 & 2.28         & \cellcolor[HTML]{CBCEFB}-  \\
\rowcolor[HTML]{EFEFEF} %\rowcolor[HTML]{E8E5E5} 
Mask-SD-VAE            & 1.21(-13.1\%) & 1.31(-4.0\%) & 5.60(+4.1\%)                         & 2.35(-3.1\%) & \cellcolor[HTML]{CBCEFB}-4.1\% \\
\rowcolor[HTML]{EFEFEF} %\rowcolor[HTML]{DAD7D7} 
ViT-VAE~\cite{han2024cra5}                & 0.92(+14.0\%) & 1.36(-7.9\%) & 4.28(+26.7\%)                        & 2.45(-7.4\%) & \cellcolor[HTML]{CBCEFB}+6.3\%  \\
\rowcolor[HTML]{C0C0C0} 
Mask-ViT-VAE           & 0.78(\textcolor{red}{+27.1\%}) & 1.29(-2.3\%) & {\color[HTML]{000000} 4.11(\textcolor{red}{+29.6\%})} & 2.41(-5.7\%) & \cellcolor[HTML]{CBCEFB}+12.2\% \\ \hline
\end{tabular}
}
\label{tab:vae}
\end{table}

% \begin{table}[t]
% \centering
% \begin{tabular}{l|c|c|c|c|c}
% \toprule
% Modalities      & AMSU-A(1e-3) & ATMS(1e-2) & HIRS(1e-3) & MHS(1e-2) & Improvement \\ \hline
% SD-VAE       &   1.07     &  1.26    &  5.84    &  2.28   & - \\ \hline
% Mask-SD-VAE  &   1.21(-13.1\%)    &  1.31 (-4.0\%)    &  5.60(+4.1\%)    &  2.35(-3.1\%)  & -4.1\% \\ \hline
% ViT-VAE       &  0.92(+14.0\%)      & 1.36 (-7.9\%)   &  4.28(+26.7\%)   &   2.45(-7.4\%) & +6.3\% \\ \hline
% Mask-ViT-VAE &   0.78(+27.1\%)     &  1.29 (-2.3\%)   &  4.11(+29.6\%)    &  2.41(-5.7\%) & +12.2\% \\ \bottomrule
% \end{tabular}
% \caption{Placeholder.}
% \end{table}

%% file: Tables/training_vae.tex
\begin{table}[!tb]
    \centering
    \caption{Hyperparameters for training the mask ViT-VAE of DAWP on the composite dataset.}
    \begin{tabular}{l|c}
    	\toprule[1.5pt]
    	Hyper-parameter & Value \\
    	\midrule\midrule
        Learning rate               & 0.0001     \\
        $\beta_1$                   & 0.9       \\
        $\beta_2$                   & 0.999     \\
        Weight decay                & 0.00001   \\
        Batch size                  & 200        \\
        Training steps              & 200000       \\
        Warm up percentage          & 10\%      \\
        Warmup learning rate        & 0.000001   \\
        Learning rate decay         & Cosine    \\
        Min learning rate           & 0.000001      \\
        KL-loss weight              & 0.000001 \\
        \bottomrule[1.5pt]
    \end{tabular}
    \label{table:vae_optimization}
\end{table}

%% file: Tables/training_aida.tex
\begin{table}[!tb]
    \centering
    \caption{Hyperparameters for training the AIDA module of DAWP on the composite dataset.}
    \begin{tabular}{l|c}
    	\toprule[1.5pt]
    	Hyper-parameter & Value \\
    	\midrule\midrule
        Learning rate               & 0.0001     \\
        $\beta_1$                   & 0.9       \\
        $\beta_2$                   & 0.999     \\
        Weight decay                & 0.00001   \\
        Batch size                  & 48        \\
        Training steps              & 200000       \\
        Warm up percentage          & 10\%      \\
        Warmup learning rate        & 0.000001   \\
        Learning rate decay         & Cosine    \\
        Min learning rate           & 0.000001      \\
        \bottomrule[1.5pt]
    \end{tabular}
    \label{table:aida_optimization}
\end{table}

%% file: Tables/training_aiwp.tex
\begin{table}[!tb]
    \centering
    \caption{Hyperparameter for training the AIWP module of DAWP on the composite dataset.}
    \begin{tabular}{l|c}
    	\toprule[1.5pt]
    	Hyper-parameter & Value \\
    	\midrule\midrule
        Learning rate               & 0.0001     \\
        $\beta_1$                   & 0.9       \\
        $\beta_2$                   & 0.999     \\
        Weight decay                & 0.00001   \\
        Batch size                  & 8        \\
        Training steps              & 200000       \\
        Warm up percentage          & 10\%      \\
        Warmup learning rate        & 0.000001   \\
        Learning rate decay         & Cosine    \\
        Min learning rate           & 0.000001      \\
        \bottomrule[1.5pt]
    \end{tabular}
    \label{table:aiwp_optimization}
\end{table}

%% file: Tables/vae_structure.tex
\begin{table}[!tb]
    \caption{
    The details of the mask ViT-VAE model on different satellite datasets. $\mathtt{Conv16\times16}$ is the 2D convolutional layer with $16\times16$ kernel. 
    The $\mathtt{FFN}$ consists of two $\mathtt{Linear}$ layers separated by a $\mathtt{GeLU}$ activation layer~\cite{hendrycks2016gaussian}. 
    The operator SamplePosterior samples a latent representation from $\mu$ and $\sigma$ as SD did~\cite{rombach2022high}.}
    \label{table:vae_structure}
    \begin{center}
        \resizebox{\textwidth}{!}{
        \begin{tabular}{l|l|c|c}
        \toprule[1.5pt]
        Module                                               & Layer                     & Resolution                        & Channels          \\
        \midrule\midrule
        Input                                               & -                         & $144\times144$                    & $c$               \\\hline     
        \multirow{3}{*}{PatchEmbed}                 & $\mathtt{Conv16\times16}$   & $9\times9$                    & $c\rightarrow768$  \\
                                                            & $\mathtt{Flatten}$   & $9\times9\rightarrow81$                    & $768$   \\
                                                            & $\mathtt{PosEmbed}$   & $81$                    & $768$
                                                            \\\hline
        \multirow{4}{*}{Trnasformer Block $\times$ 10 }                 & $\mathtt{LayerNorm}$   & $81$                    & $768$ \\
                                                            & $\mathtt{MaskAttention}$   & $81$                    & $768$              \\
                                                            & $\mathtt{LayerNorm}$    & $81$                    & $768$              \\
                                                            & $\mathtt{FFN}$    & $81$                    & $768$              \\\hline
        \multirow{6}{*}{Qauntify }                 & $\mathtt{Transformer Block}$   & $81$                    & $768$ \\
                                                    & $\mathtt{Transformer Block}$   & $81$                    & $768$ \\
                                                    & $\mathtt{Concat}$   & $81$                    & $768\rightarrow1536$ \\
                                                    & $\mathtt{Linear}$   & $81$                    & $1536\rightarrow8c$ \\
                                                    & $\mathtt{Sample Posterior}$   & $81$                    & $8c\rightarrow4c$ \\
                                                    & $\mathtt{Linear}$   & $81$                    & $4c\rightarrow768$ \\\hline
        \multirow{4}{*}{Trnasformer Block $\times$ 12}                 & $\mathtt{LayerNorm}$   & $81$                    & $768$ \\
                                                            & $\mathtt{MaskAttention}$   & $81$                    & $768$              \\
                                                            & $\mathtt{LayerNorm}$    & $81$                    & $768$              \\
                                                            & $\mathtt{FFN}$    & $81$                    & $768$              \\\hline
        \multirow{4}{*}{Out}                   & $\mathtt{Rearrange}$  & $81 \rightarrow9\times9$  & $768$  \\
                                                            & $\mathtt{Conv1\times1}$   & $9\times9$                    & $768\rightarrow256c$ \\
                                                            & $\mathtt{Rearrange}$    & $9\times9\rightarrow144\times144$          & $256c\rightarrow c$   \\
                                                            & $\mathtt{Conv3\times3}$   & $144\times144$                    & $c$ \\
                                                
        \bottomrule[1.5pt]
        \end{tabular}
        }  % end of resizebox
        \end{center}
\end{table}

%% file: Tables/mmae_structure.tex
\begin{table}[!tb]
    \caption{
    The details of the MMAE on encoded multimodal tokens within a sub-image in a time window 12. $\mathtt{Conv1\times1}$ is the 2D convolutional layer with $1\times1$ kernel. $(c_1, c_2, c_3, c_4)$ means a multimodal input list with input channels $c_1$, $c_2$, $c_3$, and $c_4$. The MaskTokens is similar to the function of random\_mask in ~\cite{he2022masked}
    , while adding the [EOS] tokens to keep the sequences from different samples the same length. The operator of PaddingTokens fills the feature map as ~\cite{he2022masked} did.
    The $\mathtt{FFN}$ consists of two $\mathtt{Linear}$ layers separated by a $\mathtt{GeLU}$ activation layer~\cite{hendrycks2016gaussian}.}
    \label{table:mmae_structure}
    \begin{center}
        \resizebox{\textwidth}{!}{
        \begin{tabular}{l|l|c|c}
        \toprule[1.5pt]
        Module                                               & Layer                     & Resolution                        & Channels          \\
        \midrule\midrule
        Multimodal Input                                               & -                         & $9\times9\times12$                    & $(c_1, c_2, c_3, c_4)$               \\\hline     
        \multirow{6}{*}{Multimodal PatchEmbed}                 & $\mathtt{Conv1\times1}$   & $9\times9\times12$                    & $(c_1, c_2, c_3, c_4)\rightarrow(768,768,768,768)$  \\
                                                            & $\mathtt{Flatten}$   & $9\times9\times12\rightarrow81\times12$                    & $(768,768,768,768)$   \\
                                                            & $\mathtt{PosEmbed}$   & $81\times12$                    & $(768,768,768,768)$ \\
                                                            & $\mathtt{TemporalEmbed}$   & $81\times12$                    & $(768,768,768,768)$ \\
                                                            & $\mathtt{Rearrange}$   & $81\times12\rightarrow3888$                    & $(768,768,768,768)\rightarrow768$
                                                            \\\hline
        Random Masking                                      & $\mathtt{MaskTokens}$                  & $3888\rightarrow128$                    & $768$               \\\hline   
        \multirow{4}{*}{Trnasformer Block $\times$ 12 }     & $\mathtt{LayerNorm}$   & $128$                    & $768$ \\
                                                            & $\mathtt{MaskAttention}$   & $128$                    & $768$              \\
                                                            & $\mathtt{LayerNorm}$    & $128$                    & $768$              \\
                                                            & $\mathtt{FFN}$    & $128$                    & $768$              \\\hline
        \multirow{6}{*}{Feature Map Filling}                & $\mathtt{Linear}$ & $128$ & $768\rightarrow512$ \\
                                                            & $\mathtt{PaddingTokens}$    & $128\rightarrow3888$                    & $512$       \\
                                                            & $\mathtt{Rearrange}$   & $3888\rightarrow81\times12\times4$                    & $512$       \\
                                                            & $\mathtt{PosEmbed}$ & $81\times12\times4$ & $512$ \\
                                                            & $\mathtt{TemporalEmbed}$ & $81\times12\times4$ & $512$ \\
                                                            & $\mathtt{Rearrange}$   & $81\times12\times4\rightarrow3888$    & $512$       \\\hline
        \multirow{4}{*}{Trnasformer Block $\times$ 8}                 & $\mathtt{LayerNorm}$   & $3888$                    & $512$ \\
                                                            & $\mathtt{Attention}$   & $3888$                    & $512$              \\
                                                            & $\mathtt{LayerNorm}$    & $3888$                    & $512$              \\
                                                            & $\mathtt{FFN}$    & $3888$                    & $512$              \\\hline
        \multirow{4}{*}{Multimodal Out}                   & $\mathtt{Rearrange}$  & $3888 \rightarrow 972$  & $(512,512,512,512)$  \\
                                                            & $\mathtt{LayerNorm}$  & $972$  & $(512,512,512,512)$  \\
                                                            & $\mathtt{Linear}$   & $972$                    & $(c_1, c_2, c_3, c_4)$ \\
                                                            &  $\mathtt{Rearrange}$ & $972\rightarrow9\times9\times12$ & $(c_1, c_2, c_3, c_4)$ \\
        \bottomrule[1.5pt]
        \end{tabular}
        }  % end of resizebox
\end{center}

\end{table}

%% file: Tables/aiwp_structure.tex
\begin{table}[!tb]
    \caption{
    The details of our AIWP module training in the assimilated space. The inputs of this module are sub-images with 8 neighbours in a multimodal way. $(c_1, c_2, c_3, c_4)$ means a multimodal input list with input channels $c_1$, $c_2$, $c_3$, and $c_4$. $Tview$ and $Sview$ indicate the temporal dimension and the spatial dimension as the sequence, respectively.
    The $\mathtt{FFNwithSwiGLU}$ consists of two $\mathtt{Linear}$ layers separated by a $\mathtt{SwiGLU}$ activation layer~\cite{shazeer2020glu}.}
    \label{table:aiwp_structure}
    \begin{center}
        \resizebox{\textwidth}{!}{
        \begin{tabular}{l|l|c|c}
        \toprule[1.5pt]
        Module                                               & Layer                     & Resolution                        & Channels          \\
        \midrule\midrule
        % Multimodal Subimg Input with Neighbor Conditions   
        Input with Conditions
        & -                         & $27\times27\times12$                    & $(c_1, c_2, c_3, c_4)$               \\\hline     
        \multirow{4}{*}{PatchEmbed}                 & $\mathtt{Concat}$ & $27\times27\times12$                    & $(c_1, c_2, c_3, c_4)\rightarrow c_{1}+c_{2}+c_{3}+c_{4}$  \\
                                                            & $\mathtt{Linear}$   & $27\times27\times12$                    & $c_{1}+c_{2}+c_{3}+c_{4}\rightarrow 768$  \\
                                                            & $\mathtt{Rearrange}$   & $27\times27\times12\rightarrow 729\times12$                    & $768$ \\
                                                            & $\mathtt{PosEmbed}$   & $729\times12$                    & $768$ \\\hline
        \multirow{10}{*}{TS Block $\times$ 12 }              & $\mathtt{T view}$   & $729\times12\rightarrow(729\times)12$                    & $768$ \\
                                                            & $\mathtt{Attention}$   & $(729\times)12$             & $768$ \\
                                                            & $\mathtt{LayerNorm}$   & $(729\times)12$             & $768$ \\
                                                            & $\mathtt{FFN with SwiGLU}$    & $(729\times)12$             & $768$ \\
                                                            & $\mathtt{LayerNorm}$   & $(729\times)12$             & $768$ \\
                                                            & $\mathtt{S view}$   & $(729\times)12\rightarrow(12\times)729$         & $768$ \\
                                                            & $\mathtt{Attention}$   & $(12\times)729$             & $768$ \\
                                                            & $\mathtt{LayerNorm}$   & $(12\times)729$             & $768$ \\
                                                            & $\mathtt{FFN with SwiGLU}$    & $(12\times)729$             & $768$ \\
                                                            & $\mathtt{LayerNorm}$   & $(12\times)729$             & $768$ \\\hline
        \multirow{3}{*}{Multimodal Out}                   & $\mathtt{Rearrange}$  & $(12\times)729\rightarrow 7\times27\times12$  & $768$  \\
                                                            & $\mathtt{LayerNorm}$  & $7\times27\times12$  & $768$  \\
                                                            & $\mathtt{Linear}$   & $972$                    & $(c_1, c_2, c_3, c_4)$ \\
        \bottomrule[1.5pt]
        \end{tabular}
        }  % end of resizebox
\end{center}
\end{table}

%% file: Tables/computation_cost.tex
\begin{table}[t]
\caption{Computation cost during inference.}
\begin{tabular}{l|c|c|c|c}
\cline{1-5}
                                   & Inference time(ms) & Parameters(MB) & Memory(MB) & Batch size(per GPU) \\ \hline
\multicolumn{1}{l|}{Mask-ViT-VAE} & 53                 & 96             & 7262       & 50                  \\ 
\multicolumn{1}{l|}{AIDA}         & 310                & 105            & 18242      & 12                  \\ 
\multicolumn{1}{l|}{AIWP}         & 491                & 216            & 47134      & 2                   \\ \hline
\end{tabular}
\label{tab:comp_cost}
\end{table}